\documentclass[journal]{IEEEtran}
\usepackage{cite}

\ifCLASSINFOpdf
  \usepackage[pdftex]{graphicx}
	\graphicspath{{fig/}}
\else
\fi
\usepackage{amssymb,amsmath}

\usepackage{algorithm}
\usepackage{algorithmic}

\usepackage{subfigure}
\usepackage{booktabs} 
\usepackage{multirow}
\usepackage{color}
\usepackage{bbm}
\usepackage[switch,columnwise]{lineno}
\usepackage{mathtools}
\DeclarePairedDelimiter{\ceil}{\lceil}{\rceil}
\usepackage{textcomp}

\newcommand{\etal}{\textit{et al.}}

\begin{document}
%
\title{Detecting Colorized Images via Convolutional Neural Networks: Toward High Accuracy \\ and Good Generalization}

\author{Weize~Quan,
        ~Dong-Ming~Yan,
        ~Kai~Wang,
        ~Xiaopeng~Zhang
        and Denis~Pellerin
\thanks{W. Quan is with the National Laboratory of Pattern Recognition, Institute of Automation Chinese Academy of Sciences, Beijing 100190, China; the University of Chinese Academy of Sciences, Beijing 100049, China; and Univ. Grenoble Alpes, CNRS, Grenoble INP, GIPSA-lab, 38000 Grenoble, France (e-mail: qweizework@gmail.com).}
\thanks{D.-M. Yan and X. Zhang are with the National Laboratory of Pattern Recognition, Institute of Automation Chinese Academy of Sciences, Beijing 100190, China; and the University of Chinese Academy of Sciences, Beijing 100049, China (e-mail: yandongming@gmail.com; xiaopeng.zhang@ia.ac.cn).}
\thanks{K. Wang and D. Pellerin are with Univ. Grenoble Alpes, CNRS, Grenoble INP, GIPSA-lab, 38000 Grenoble, France (e-mail: kai.wang@gipsa-lab.grenoble-inp.fr; denis.pellerin@gipsa-lab.grenoble-inp.fr).}}

\maketitle

\begin{abstract}
Image colorization achieves more and more realistic results with the increasing computation power of recent deep learning techniques. It becomes more difficult to identify the fake colorized images by human eyes. In this work, we propose a novel forensic method to distinguish between natural images (NIs) and colorized images (CIs) based on convolutional neural network (CNN). Our method is able to achieve high classification accuracy and cope with the challenging scenario of blind detection, \textit{i.e.}, no training sample is available from ``unknown'' colorization algorithm that we may encounter during the testing phase. This blind detection performance can be regarded as a generalization performance. First, we design and implement a base network, which can attain better performance in terms of classification accuracy and generalization (in most cases) compared with state-of-the-art methods. Furthermore, we design a new branch, which analyzes smaller regions of extracted features, and insert it into the above base network. Consequently, our network can not only improve the classification accuracy, but also enhance the generalization in the vast majority of cases. To further improve the performance of blind detection, we propose to automatically construct negative samples through linear interpolation of paired natural and colorized images. Then, we progressively insert these negative samples into the original training dataset and continue to train the network. Experimental results demonstrate that our method can achieve stable and high generalization performance when tested against different state-of-the-art colorization algorithms.
\end{abstract}


\IEEEpeerreviewmaketitle

\section{Introduction}

\begin{figure}
\centering
\includegraphics[width=1.00\linewidth]{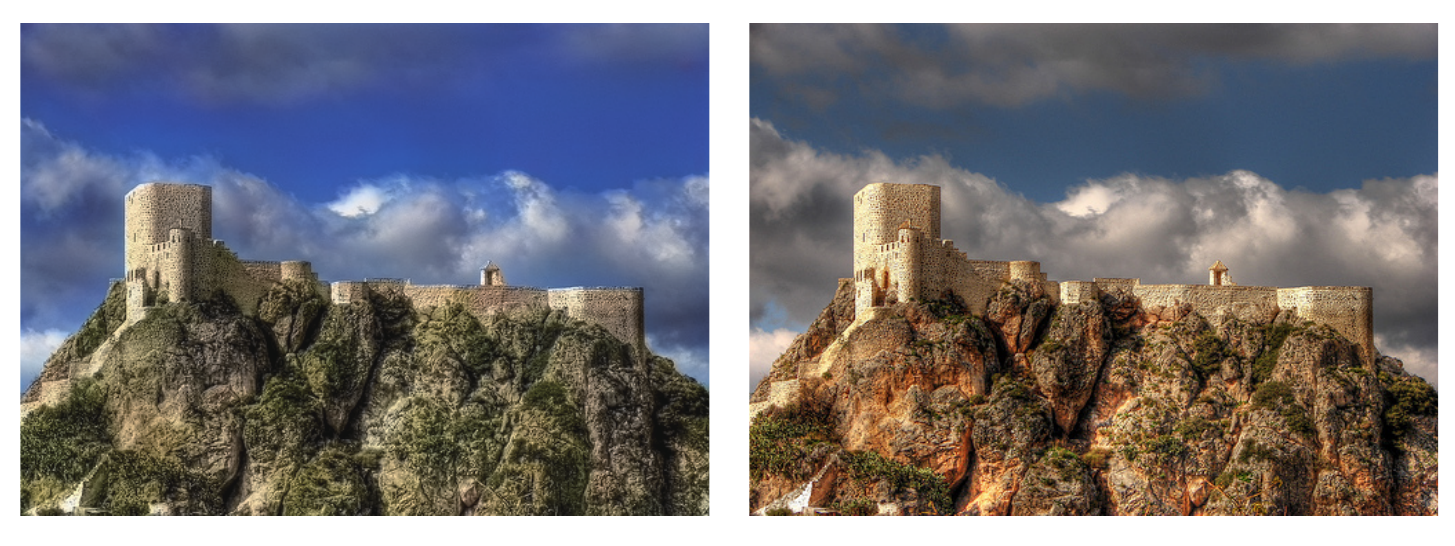}
\caption{Pair of images. The left one is a colorized image generated by the colorization method proposed in~\cite{iizuka_let_2016}, and the right one is a natural image taken from ImageNet~\cite{ImageNet2009}.}
\label{fig:color}
\end{figure}

\IEEEPARstart{W}{ith} the increasing popularity and sophistication of image editing technologies, it is now relatively easy to create edited images that are visually plausible. For example, current advanced colorization algorithms, more or less leveraging the powerful capacity of deep neural networks, can automatically colorize the grayscale images to obtain the high-quality color images. Fig.~\ref{fig:color} shows a pair of images, the right one is the original color image used for comparison, and the left one is a colorized image produced by a fully automatic colorization algorithm~\cite{iizuka_let_2016} that takes the grayscale version of the right one as input. Obviously, it is difficult to distinguish which one is colorized image by naked human eyes. Although this technique brings convenience to people's live, it may also be maliciously used and potentially lead to security issues, such as confounding object recognition or scene understanding~\cite{guo_fake_2018}. Therefore, distinguishing between natural images (NIs) and colorized images (CIs) has become an important research problem in image forensics.

Very recently, Guo \etal~\cite{guo_fake_2018} first proposed two approaches to solve this new forensic problem. On the basis of the statistical difference between NIs and CIs in the hue, saturation, dark, and bright channels, two methods, namely, histogram-based and Fisher-encoding-based, were designed to catch the difference. After having obtained the discriminant feature vectors, they trained the support vector machine (SVM) classifiers to identify fake colorized images. In fact, the classification performance of their methods still has some space for improvement. Furthermore, in the challenging scenario of \emph{blind detection}, \textit{i.e.}, no training sample is available from ``unknown'' colorization method that we may encounter during the testing phase of forensic detectors, the performance of their methods in general decreases. Hereafter, we call this blind detection performance as \emph{generalization} performance. In the meanwhile, although not being very rigorous, we choose to use the term ``classification accuracy/performance'' to indicate the detection performance on testing data in which CIs are generated by a same colorization method known by the training procedure.

Nowadays, convolutional neural network (CNN) has obtained obvious performance improvement compared with traditional handcrafted-feature-based methods, not only in computer vision and pattern recognition~\cite{AlexNet2012,RCNN2014,DenseNet2017}, but also in multimedia security~\cite{qian_deep_2015,chen_median_2015,ye_deep_2017,quan_distinguishing_2018,bayar_deep_2018}. A well-known reason is that it can automatically extract useful information from (complex) data and thus has powerful learning capacity. In addition, its unified optimization framework, \textit{i.e.}, in the ``end-to-end'' manner, may be superior than the multi-step pipeline of conventional methods which often have separate stages of extracting handcrafted features (somehow reflecting human prior knowledge on the problem) and training classifiers. In this work, we propose a CNN-based method to identify colorized images. Specifically, we propose two ways to improve the forensic performance, especially the generalization capability. We design in the first place a base network and then improve its architecture, with the objective to obtain better performance in terms of classification accuracy and generalization. Afterwards, in order to better cope with the challenging scenario of blind detection, we introduce a simple yet effective method, namely, inserting additional auto-constructed negative samples into the original training dataset and then carrying out enhanced training of the network for a better generalization performance.

Our main contributions are summarized below:
\begin{itemize}
\item We design and implement a base ``end-to-end'' deep model based on CNN to identify NIs and CIs, which obtains better classification accuracy and generalization capability (in most cases) compared with state-of-the-art methods~\cite{guo_fake_2018}. We also consider and compare three different design choices about the activation of the network's first layer.
\item We improve the original base network via inserting a new branch, which analyzes the smaller regions of the extracted features of the first layer, to enrich the learned features and enhance the discrimination capacity of network. This enhanced network can not only increase the classification accuracy, but also improve the generalization performance in the vast majority of cases.
\item We introduce a simple yet effective method to further improve the generalization performance of the proposed network. In practice, we construct negative samples via linear interpolation of paired natural and colorized images available in the training dataset, and iteratively add them into the original training dataset for additional and enhanced CNN training. This procedure is fully automatic, and can allow us to obtain stable and high generalization performance when conducting tests against colorization algorithms that are ``unknown'' during the training stage.
\end{itemize}

The rest of this paper is organized as follows. Section~\ref{sec:related work} reviews relevant existing work. Section~\ref{sec:proposed framework} discusses the motivation of every step of our work, and presents the details of the proposed method. Section~\ref{sec:experimental_results} reports the performance evaluations for our method and comprehensive comparisons with state-of-the-art methods. Section~\ref{sec:conclusion} draws the conclusions and proposes some future working directions.

\section{Related Work}
\label{sec:related work}

\subsection{Colorized Image and Its Identification}
Image colorization adds color to a monochrome image and obtains a realistic color image. Existing colorization algorithms mainly consist of three categories: scribble-based~\cite{levin_colorization_2004,luan_natural_2007,xu_efficient_2009,chen_manifold_2012,pang_image_2013}, reference-based~\cite{welsh_transferring_2002,irony_colorization_2005,gupta_image_2012}, and fully automatic~\cite{cheng_deep_2015,iizuka_let_2016,larsson_learning_2016,zhang_colorful_2016} approaches.

Scribble-based methods require user-specific scribbles and propagate the color information to the whole grayscale images. This kind of method is usually accompanied by trail and error to obtain satisfactory results, and thus is rather time-consuming. Reference-based (or example-based) approaches mainly exploit the color information of a reference image that is (semantically) similar to the input grayscale image. The core idea is to model a matching relationship between these two types of images. However, the selection of suitable reference image may be burdensome.

In contrast, recently researchers have developed fully automatic methods that do not need user interaction or example color images, and that are usually working in the data-driven manner. Cheng \etal~\cite{cheng_deep_2015} proposed the first deep neural network based image colorization method. Their method performed pixel-wise prediction, however the input of deep model was pre-extracted handcrafted features. Iizuka \etal~\cite{iizuka_let_2016} proposed a novel fully ``end-to-end'' network for the task of image colorization. The input was a grayscale image and its output was the chrominance, which was combined with the input image to produce the color image. Their network jointly learned global and local features from an image, and at the same time, they also exploited classification labels of the grayscale images to improve the performance. Different from previous methods, Larsson \etal~\cite{larsson_learning_2016} proposed a deep model that predicted a color histogram, instead of a single color value, at every image pixel. Zhang \etal~\cite{zhang_colorful_2016} took into account the nature of uncertainty of this colorization task and introduced class-rebalancing method to increase the diversity of color of resultant image. These CNN-based methods lead to the very high visual quality of colorized images, often plausible enough to deceive the human perception.

As shown in Fig.~\ref{fig:color}, visually realistic colorized image (the left one), which is generated by the state-of-the-art colorization algorithm~\cite{iizuka_let_2016}, is difficult to distinguish compared with corresponding natural image (the right one). Very recently, Guo \etal~\cite{guo_fake_2018} first proposed handcrafted-feature-based methods to detect the fake colorized images. On the basis of the observation that the colorized images tend to possess less saturated colors, they analyzed the statistical difference between NIs and CIs in the hue and saturation channels. In addition, they also found that there are differences in certain image priors. In practice, they exploited the extreme channels prior (ECP)~\cite{yan_image_2017}, \textit{i.e}, the dark channel prior (DCP)~\cite{he_single_2011} and the bright channel prior (BCP). They proposed two approaches, \textit{i.e.}, histogram-based and Fisher-encoding-based, to extract statistical features, and then trained SVMs for classification. We believe that this new and important forensic problem deserves further studies because the results shown in the pioneer work~\cite{guo_fake_2018} could be improved in terms of classification accuracy and generalization performance~--~both are important metrics for moving forensic algorithms towards practical applications. The same as in Guo \etal's work~\cite{guo_fake_2018}, in our study, we also consider high-quality colorized images generated by three state-of-the-art colorization algorithms, hereafter denoted respectively by Ma~\cite{larsson_learning_2016}, Mb~\cite{zhang_colorful_2016}, and Mc~\cite{iizuka_let_2016}.

\subsection{CNN for Multimedia Security}
Inspired by the notable success of CNN, in the multimedia security community, a number of researchers have used CNN for image forensics~\cite{chen_median_2015,tuama_camera_2016,bondi_first_2017,barni_aligned_2017,amerini_localization_2017,nicolas_cnn_2017,quan_distinguishing_2018,bunk_detect_2017,bappy_exploit_2017,bayar_deep_2018} and steganalysis~\cite{qian_deep_2015,xu_structural_2016,pibre_deep_2016,chen_jpeg_2017,ye_deep_2017}.

Concerning CNN-based image forensics, different research problems have been considered. Chen \etal~\cite{chen_median_2015} first proposed to use CNN to detect median filtering, and obtained significant performance improvement compared with traditional methods. Tuama \etal~\cite{tuama_camera_2016} and Bondi \etal~\cite{bondi_first_2017} utilized CNN to accomplish the task of source camera identification. This powerful tool was also employed to distinguish between natural and computer graphics images~\cite{nicolas_cnn_2017,quan_distinguishing_2018}, and to detect image forgery~\cite{bunk_detect_2017,bappy_exploit_2017}. In addition, Bayar \etal~\cite{bayar_deep_2018} developed a so-called constrained convolutional neural network to solve general purpose image manipulation detection problem.

Most of previous CNN-based methods mentioned above use conventional single-stream networks to complete their tasks~\cite{chen_median_2015,qian_deep_2015,tuama_camera_2016,xu_structural_2016,bondi_first_2017,bappy_exploit_2017,quan_distinguishing_2018,bayar_deep_2018}. Different from this conventional design, other design choices have been considered, for example injecting additional knowledge to CNN~\cite{chen_jpeg_2017} and utilizing multi-stream inputs (\textit{i.e.}, multiple representations of the same input image in different domains)~\cite{barni_aligned_2017,amerini_localization_2017}. Chen \etal~\cite{chen_jpeg_2017} introduced JPEG-phase knowledge into the CNN architecture to detect modern JPEG steganography. Barni \etal~\cite{barni_aligned_2017} designed the CNN-based model for aligned and nonaligned double JPEG compression detection. Their networks took three inputs: original images, noise residuals, and discrete cosine transform (DCT) histograms (with an additional sub-network to compute DCT histograms), respectively. They fused the output of DCT-based CNN and noise-based CNN as feature vector, and then trained a random forest to improve the accuracy in the mixed case of aligned and misaligned compression. Different from this ``hard'' fusing strategy, Amerini \etal~\cite{amerini_localization_2017} fused deep features of two networks with different inputs, \textit{i.e.}, original images and DCT histograms, using fully connected layer, and thus the two-stream network can be trained in an ``end-to-end'' manner. In our work, we propose a two-branch network. Unlike previous networks mentioned above, the input of our network is only the image under forensic examination, without any additional knowledge or a different representation of the image. In addition, our feature fusion locates at the middle of network, and there are several convolutional layers after feature fusion to further learn hierarchical and discriminative representation for detecting colorized images.

At last, to the best of our knowledge, there is no existing work that considers the ``generalization'' capability yet for CNN-based image forensics. In fact, this is a highly challenging scenario because no training samples of the ``unknown'' colorization algorithms are available. In other words, we want the trained network to be able to successfully detect colorized images generated by new colorization methods that remain unknown during the training of CNN. In this work, we solve this challenging generalization problem through a simple yet effective approach, \textit{i.e.}, inserting additional negative samples that are automatically constructed from available training samples, in order to carry out an enhanced training of CNN.

\section{Proposed Framework}
\label{sec:proposed framework}

\begin{figure*}
\centering
\includegraphics[width=0.79\linewidth]{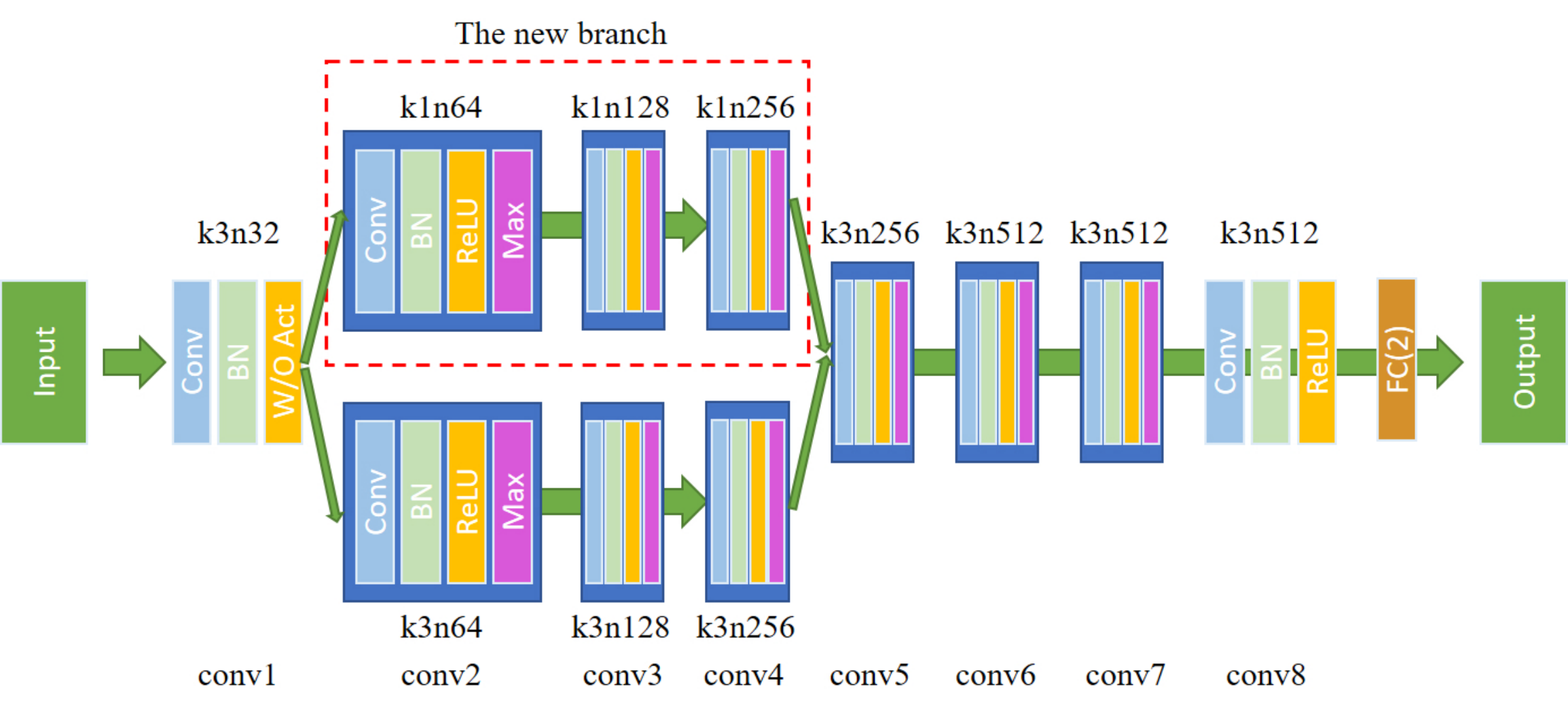}
\vspace{-2mm}
\caption{Architecture of our networks named respectively by BaseNet (architecture excluding the part within the red dotted rectangle, Section \ref{subsec:BaseNet}) and DecNet (whole architecture, Section \ref{subsec:DecNet}). The network input is a $256\times256$ RGB image, and output is the class scores. For each convolutional layer, k is the kernel size and n is the number of feature maps. The two-branch outputs of conv4 have same size and are directly concatenated as the input of conv5. ``W/O Act'' means ``with or without activation'', and ``FC(2)'' stands for a fully-connected classifier layer with a 2-dimensional output of class scores.}
\label{fig:network}
\end{figure*}

\subsection{Motivation}
Our study is inspired by Guo \etal's work~\cite{guo_fake_2018}, where they first proposed histogram-based and Fisher-encoding-based fake colorized image detection methods, and obtained decent performance. In fact, these two handcrafted features to some extent are based on the prior knowledge observed from data, and thus may be the non-optimal discriminant features for this complex identification task. The classification accuracy shown in~\cite{guo_fake_2018} can support this point as well. Furthermore, the generalization performance could be further improved as discussed in~\cite{guo_fake_2018}. More specifically, the forensic performance of their methods sometimes drops when the training images and the testing images are produced by different colorization algorithms. Therefore, an ``end-to-end'' framework based on CNN could be a good solution to automatically learn informative and generic characteristics between natural and colorized images. In our approach, we consider two aspects: (1) designing a suitable CNN architecture to learn discriminative and enriched features for this forensic problem with good classification accuracy and generalization performance, and (2) constructing additional training data, \textit{i.e.}, the so-called negative samples, to obtain an appropriate decision boundary for this classification problem and thus further improve the generalization capability of our network.

\subsection{Our Network - Base Architecture}
\label{subsec:BaseNet}
For this forensic problem, we first design and implement a base CNN. Except for the components enclosed by red dotted rectangle in Fig.~\ref{fig:network}, the remaining part is the proposed base network (called ``BaseNet'') with a conventional single-branch structure. Our network consists of 8 convolutional layers and a fully-connected classifier layer (in total 9-layer deep). Inspired by the recent network designs of computer vision tasks~\cite{ResNet2016,ResNeXt2017,DenseNet2017}, our network ends with a 2-way fully-connected layer instead of traditional stacked multi-layer perceptrons and thus has less parameters. The input of BaseNet is an RGB image. After the first layer (conv1), it is expected that much useful information is extracted from the original input image. The next three layers (conv2-4) are designed to analyze the extracted features of the first layer. Then a somehow high-level abstraction and reasoning is applied via the remaining layers (conv5-8). Finally, the 2-dimensional score vector of the class label is output by a fully-connected classifier [FC(2) with (2) standing for output dimension]. All convolutional kernel sizes in BaseNet are $3\times3$. For conv1-7, each convolutional layer (Conv) is with the zero-padding of 1, which ensures that the input and output of Conv have the same size. The loss function of our network is cross entropy loss, which is most commonly used for classification tasks. Given the training dataset of images $x_i$, each associated with a label $y_i$, where $i=1,2,...,N$ and $y_i\in\lbrace0,1\rbrace$ (0: CI and 1: NI), the loss function can be described as:
\begin{equation}
L = -\frac{1}{N}\sum_{i} log(\frac{e^{c_{y_i}}}{\sum_{j}e^{c_j}}),
\end{equation}
where $c_j $ means the $j$-th element of the class score vector $c$.

In our network, each Conv is equipped with the batch normalization (BN) layer. BN~\cite{BN2015} explicitly forces the output of Conv to take on a unit Gaussian distribution. At the same time, a pair of parameters of \textit{shift} and \textit{scale} is applied to guarantee that the transformation can represent the identity transform. This layer increases the stability of the network training and reduces the potential overfitting due to its slight regularization effects. All max-pooling layers (Max) \cite{Maxpool1988} in BaseNet have the same kernel size of $3\times3$ and a stride of 2. Max-pooling reports the maximum output within a local window of feature maps, and essentially is a down-sampling operation. This operation brings two benefits: reducing the number of parameters within the model by decreasing the spatial size of processed feature maps; and making the representation approximately invariant to small translations~\cite{goodfellow_deep_2016}. Many multimedia security researchers argue that the extracted low-level features of the first layer are crucial for the success of their tasks~\cite{chen_median_2015,xu_structural_2016,bayar_deep_2018}. In this work, we pay attention to the activation function of the first layer, and we consider and compare three different design choices (reflected by the box of ``W/O Act'' in Fig.~\ref{fig:network}): without activation, with rectified linear unit (ReLU) activation~\cite{ReLU2010}, and with hyperbolic tangent (TanH) activation. The input-output relation of ReLU activation is $f(x) = \max(0,x)$, and that of TanH is $\tanh(x) = \frac{e^x - e^{-x}} {e^x + e^{-x}}$.

Compared with state-of-the-art methods using handcrafted features~\cite{guo_fake_2018}, this BaseNet already obtains better classification accuracy and generalization performance (in most cases), and detailed results are given in Section~\ref{subsec:state}.

\begin{figure*}
\centering
\includegraphics[width=0.78\linewidth]{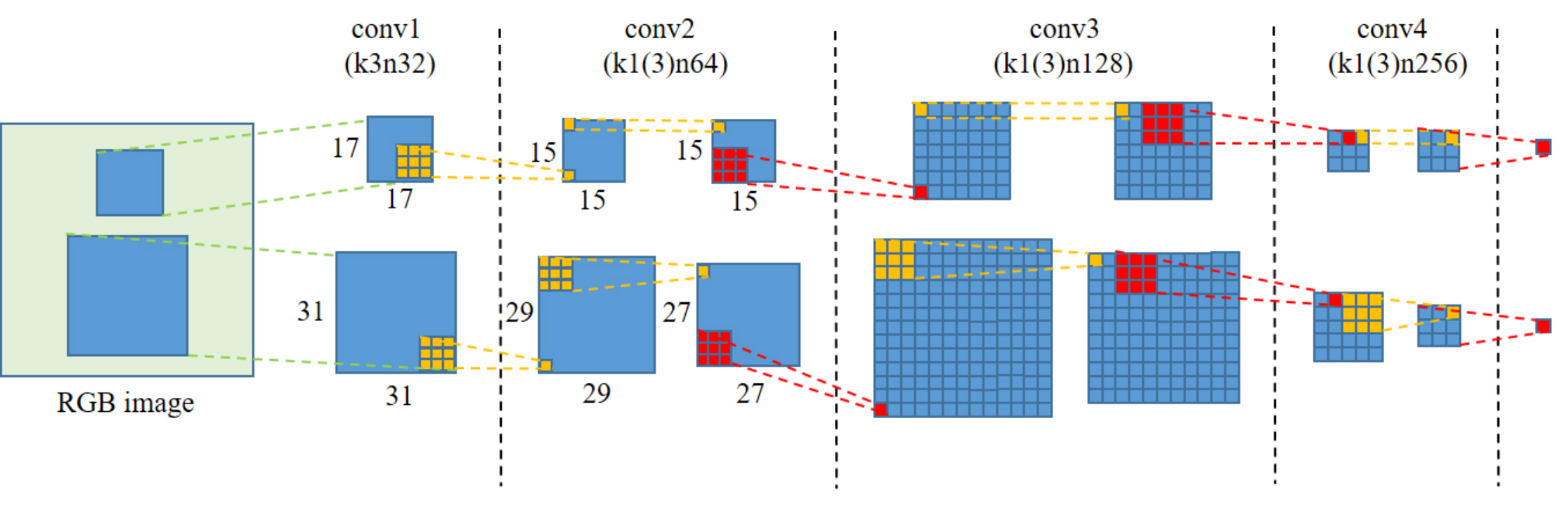}
\vspace{-2mm}
\caption{The local region sizes in the original image space ``seen'' by a neuron at output of two branches, respectively. Note that, this neuron locates at the concatenation stage of feature maps of conv4. Here, four columns (from ``conv1'' to ``conv4'') correspond to the first four layers of our network shown in Fig.~\ref{fig:network}, and we explicitly illustrate this correspondence, for example, ``conv1(k3n32)''. A blue square stands for a feature map, and the numbers close to it denote its size. A group of two yellow squares stands for convolutional operation, and a group of two red squares stands for max-pooling.}
\label{fig:twobranch}
\end{figure*}

\subsection{Our Network - Enhanced Architecture}
\label{subsec:DecNet}
In order to enhance the learning capacity of the BaseNet, we improve its architecture and the corresponding inspiration is borrowed from ensemble learning. Ensemble learning combines multiple predictions of a set of individually trained classifiers, and then gives the final decision~\cite{opitz_popular_1999}. Empirically, more variety among the base classifiers makes the ensemble more powerful~\cite{dietterich_ensemble_2000,rokach_ensemble_2010}. In our work, loosely speaking, we try to apply this idea by slightly adjusting the base network's architecture. Practically, we design a new branch which is different from the base network, and insert it in the middle of BaseNet to jointly analyze the extracted features of the first layer (conv1) from a multi-scale perspective. This enhanced network is denoted by DecNet (Detection colorization Network). The new branch is highlighted by red dotted rectangle in Fig.~\ref{fig:network}. The convolutional kernels in this new branch have the same size, \textit{i.e.}, $1\times1$, which is different from the $3\times3$ kernel of the other branch. For the Convs with same position in two branches (conv2-4), the sizes of their outputs are same because the former uses $1\times1$ kernel and the latter uses $3\times3$ kernel with zero-padding of 1. In the meantime, the settings of Max in these two branches are also consistent. Hence, the analysis results of the two branches have the same size and can be directly concatenated as the input of the fifth layer (\textit{i.e.}, conv5) of DecNet.

Due to different architectures of two branches, \textit{i.e.}, different kernel sizes, a neuron in the output of these two branches (\textit{i.e.}, the output of conv4) corresponds to regions of different sizes in the input image space. This difference is analyzed and shown in Fig.~\ref{fig:twobranch}. The region size ``seen'' by a neuron at the output of conv4 of the new branch is $17\times17$ (the top row of Fig.~\ref{fig:twobranch}), which is almost quarter of that of the base network (the corresponding region size is $31\times31$, as shown in the bottom of Fig.~\ref{fig:twobranch}), because the new branch has smaller convolutional kernel size. Therefore, the difference between two branches in terms of local region size ``seen'' in the original image space can introduce some level of variety into the process of feature analysis (conv2-4). Then, we utilize several convolutional layers (conv5-8) to efficiently fuse the analysis results of two branches, intending to make good use of this potential variety. Consequently, this enhanced network further increases the classification accuracy and improves the generalization performance in the vast majority of cases. Quantitative results are reported in Section~\ref{subsec:architecture}.

\subsection{Negative Sample Insertion}
\label{subsec:neg}

\begin{figure*}
\centering
\subfigure[]{
\includegraphics[width=0.35\textwidth]{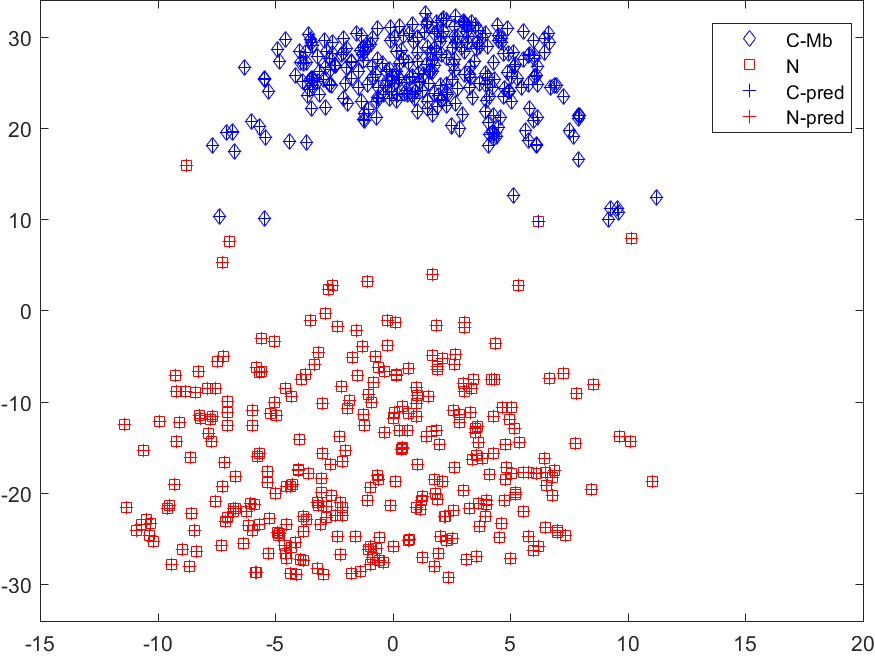}
}
\subfigure[]{
\includegraphics[width=0.35\textwidth]{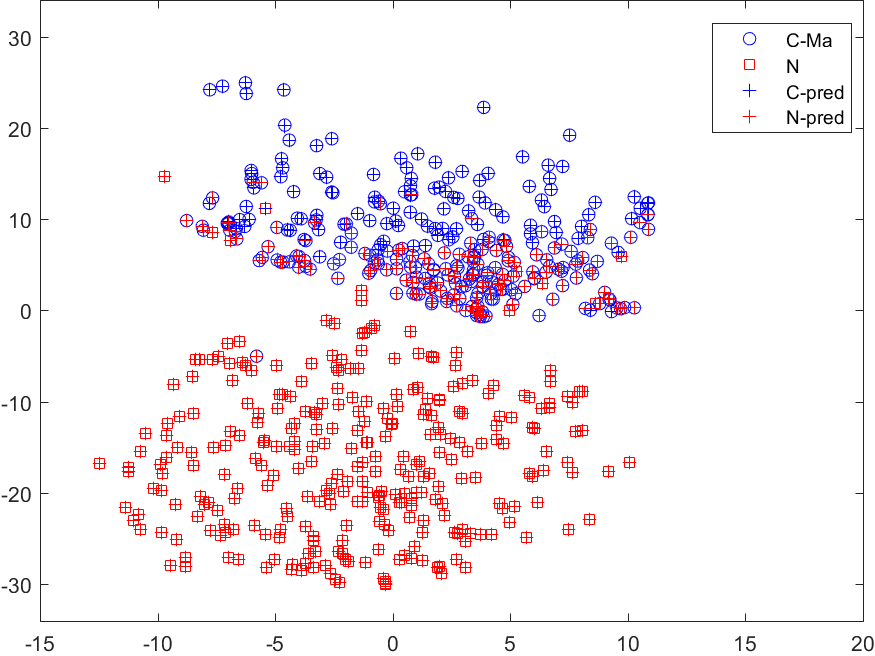}
}
\subfigure[]{
\includegraphics[width=0.35\textwidth]{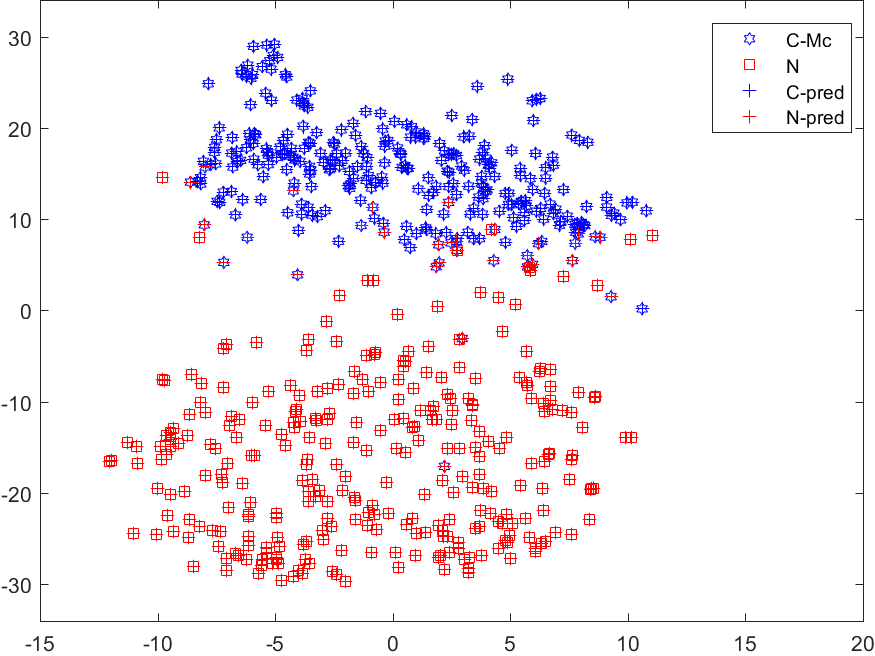}
}
\subfigure[]{
\includegraphics[width=0.35\textwidth]{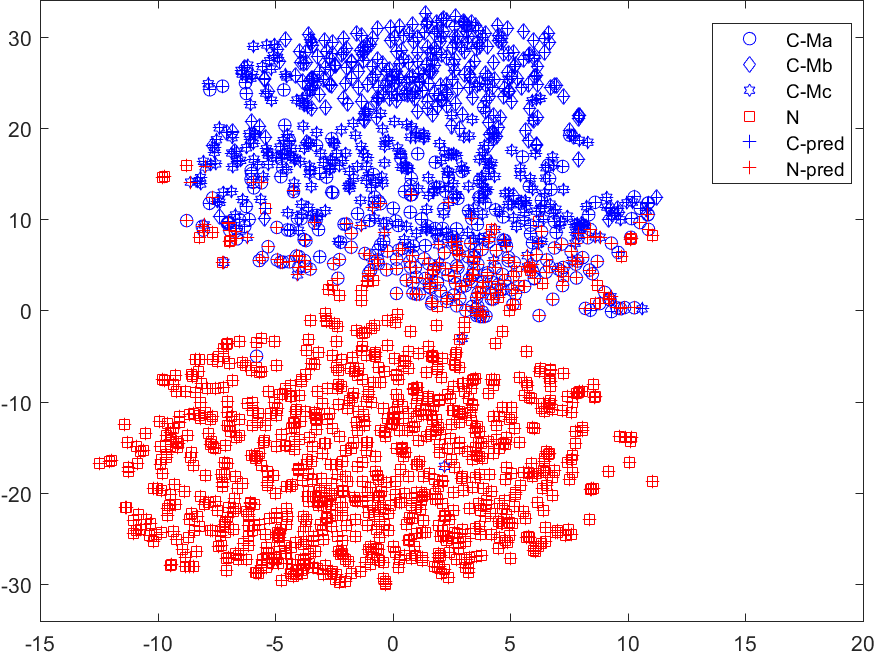}
}
\caption{The deep feature visualization with t-SNE~\cite{TSNE2008}. The model is trained on the original dataset where CIs are generated by Mb. ``C'' means colorized images and ``N'' means natural images. ``C-X'' means the colorized images produced by X colorization method, for example, ``C-Ma'' corresponds to CIs generated by Ma colorization algorithm. ``Y-pred'' means that the predicted label of CNN is Y. The network is DecNet using TanH in the first layer. We randomly select 900 natural images from validation dataset splitting them into three equal subsets of 300 images, and then we construct corresponding colorized images using Ma, Mb, and Mc for every 300 images. The deep feature is the output of conv8, and the dimension is 512. In addition, (d) is the combination of (a), (b), and (c).}
\label{fig:boundary_o}
\end{figure*}

According to our observation, there is a certain degree of performance deviation in the challenging blind detection scenario, not only for traditional handcrafted-feature-based methods~\cite{guo_fake_2018}, but also for our CNN-based approach, although the latter has better performance. In details, for a traditional or CNN-based model trained on dataset constructed by one specific colorization algorithm, the test performance on datasets constructed by other colorization algorithms is sometimes rather limited for colorized images. The possible reason of this performance drop is that colorized images produced by a specific colorization algorithm tend to be equipped with a particular internal property, but CIs of different colorization algorithms are very likely to have different properties.

To clearly illustrate the encountered problem with an example, we train the DecNet on the dataset constructed by colorization method Mb~\cite{zhang_colorful_2016}, and test on the datasets constructed by Ma~\cite{larsson_learning_2016} and Mc~\cite{iizuka_let_2016}, respectively. It should be noted that Ma and Mc are the ``unknown'' colorization algorithms, and thus the corresponding samples of Ma and Mc are not used in the training process. We use t-distributed stochastic neighbor embedding (t-SNE)~\cite{TSNE2008} to project the high-dimensional deep features (the output of conv8 of DecNet, and its dimension is 512) of testing data constructed by above three colorization methods onto the two-dimensional map, and detailed visualization results are shown in Fig.~\ref{fig:boundary_o}. Comparing Fig.~\ref{fig:boundary_o}(a), (b) and (c), we find that the distributions of NIs (red squares) are relatively stable with a rather high ``intra-class'' variation, which is somehow expected; in the meanwhile, CIs (blue symbols) are more tightly clustered for each colorization algorithm but their locations change a lot for different methods [please compare the CIs in (a), (b) and (c)]. This is reasonable because the different colorization methods tend to have not exactly the same internal characteristics and hence the corresponding CIs have different locations in the feature space. When the features of CIs produced by ``unknown'' colorization algorithms (here Ma and Mc whose samples are not used for training) are near the decision boundary of the CNN (which is trained by using NIs and CIs produced by a ``known'' colorization algorithm, here Mb), and at the same time the decision boundary is relatively close to colorized images, there are high probabilities to misclassify the ``unknown'' CIs. For instance, many CIs in Fig.~\ref{fig:boundary_o}(b) (blue circles with red + in the figure) are wrongly predicted as NIs.

We would like to find a simple yet effective method to solve the encountered problem. The idea is that we make use of the available training samples (and only these samples) to construct an appropriate decision boundary which can lead to better generalization performance. A feasible and intuitive solution is to add negative samples (with same labels as CIs) near the initial decision boundary of the CNN, so as to make the CNN be more ``strict'' about the predictions of CIs and somehow push the classification boundary towards NIs. As such, it is expected that the ``unknown'' CIs located close to the initial decision boundary [\textit{e.g.}, those shown in Fig.~\ref{fig:boundary_o}(b)] have more chance to be correctly classified with the new classification boundary which would be closer to NIs. More precisely, we construct negative sample through linear interpolation between \emph{paired} NI and CI which share the same grayscale version and only differ in chrominance components. The corresponding formulation is shown below:
\begin{equation}
I_{NS} = \alpha \cdot I_{N} + (1 - \alpha) \cdot I_{C},
\label{equ:ng}
\end{equation}
where $I_{NS}$ is the negative sample, $I_{N}$ is the natural image, $I_{C}$ is the corresponding colorized image, and $\alpha \in \lbrace0.1, 0.2, 0.3, 0.4\rbrace$ is the interpolation factor. This actually makes sense, as negative samples are in fact forensically negative (\textit{i.e.}, considered as CIs), especially for our chosen weight values among $\lbrace0.1, 0.2, 0.3, 0.4\rbrace$ (\textit{i.e.}, negative samples are closer to CIs than NIs). When $\alpha$ increases, the negative samples are progressively getting closer to the natural images and it is expected that the decision boundary is further moving towards NIs after enhanced training.

As analyzed above, adding negative samples and conducting additional training will push the classification boundary towards NIs. Thus, the classification accuracy on the NIs will gradually decrease as more and more negative samples are inserted. The classification accuracy of network on validation dataset also slightly decreases because the CIs are almost all correctly classified and this accuracy mainly depends on the classification accuracy on the NIs. However, in the meanwhile the CIs constructed by ``unknown'' colorization algorithms are expected to be classified more correctly, implying a better generalization capability. Obviously, there is a trade-off between the classification accuracy (on data similar to the training samples) and generalization performance (mainly on ``unknown'' CIs) for our network. Therefore, without being able to directly measure the generalization during training of network, we consider the classification accuracy on NIs (on the so-called \emph{natural validation dataset} $\mathcal{V}$) as a measure to select the final model in the process of additional training with negative sample insertion. In our work, we design a threshold-based model selection criterion. This threshold ($\theta$) essentially determines the degree of final classification accuracy that can be accepted by user or current task. Generally speaking, larger $\theta$ means that the selected model has less high classification accuracy, but better generalization performance. Basically, we set $\theta = \beta \cdot error\_rate$, where $\beta$ is a user defined parameter and $error\_rate$ is the classification error rate (in $\%$, measured on natural validation dataset $\mathcal{V}$) of the CNN model trained with the original training dataset $\mathcal{D}$ before negative sample insertion. This criterion simply defines the maximum tolerable value of the relative increase of error rate on $\mathcal{V}$ induced by enhanced training. In our experiments, we set $\beta = 2$. One exception is that when $error\_rate$ is very small (less than $1\%$), we set $\theta = 2\%$, meaning that we can slightly relax the constraint on classification error rate to obtain relatively large improvement of generalization performance.

Algorithm~\ref{alg:ng_insertion} illustrates the training process with negative sample insertion. It is worth noting that we only use CIs of a ``known'' colorization method but in a better way to construct a more appropriate decision boundary. In our experiments, this insertion is an iterative process with four iterations, \textit{i.e.}, the $\alpha$ is increased from 0.1 to 0.4 with step of 0.1. Given a CNN model $\mathcal{M}$ trained by using original dataset $\mathcal{D}$, and some basic settings for CNN training, such as initial learning rate $lr^0$ and $S$ epochs for each insertion, we first compute $error\_rate$ on $\mathcal{V}$ and then the threshold $\theta$, which are used for final model selection. For each round of negative sample insertion, we construct negative samples and insert them into the dataset $\mathcal{D}$. Then, we update the parameters of model $\mathcal{M}$ using new training dataset, and compute the error rate on $\mathcal{V}$ starting from the second half of training process (\textit{i.e.}, from $\ceil[\big]{\frac{S}{2}}$-th epoch for each insertion, where $\ceil[\big]{.}$ is the integer ceiling operator),  because from that time the model becomes relatively stable. After each insertion, we test the negative samples produced by previous iteration. If a negative sample is misclassified, \textit{i.e.}, the predicted label is NI and not consistent with its ground-truth label, then we stop using the corresponding pair to construct negative sample (\textit{i.e.}, we remove corresponding pair from $\mathcal{P}$ as described in line 9 of Algorithm~\ref{alg:ng_insertion}). In fact, this operation can slightly reduce the amount of negative samples, and does not weaken the performance of our network. After four iterations of insertion, we select the final CNN model. It is worth mentioning that when $\alpha \geqslant 0.5$, the negative samples will be close to NIs, and this is likely to have more impact on the classification of NIs. We take a conservative and experimentally effective approach, \textit{i.e.}, stopping the negative sample insertion process after four iterations.

\begin{algorithm}[t]
\small
    \textbf{Input:} $\mathcal{M}$, $lr^0$, $S$, $\mathcal{V}$, $\mathcal{D}$ and the set of corresponding natural and colorized image pairs $\mathcal{P}$ constructed from $\mathcal{D}$. \\
    \textbf{Output:} final model after enhanced training. \\
    \textbf{Initialization:} current learning rate $lr = lr^0$, negative samples $\mathcal{N} = \emptyset$, set of error rates on $\mathcal{V}$ of candidate CNN models $\mathcal{R} = \emptyset$.
    \begin{algorithmic}[1]
        \STATE compute $error\_rate$ of $\mathcal{M}$.
        \STATE compute $\theta$.
        \FORALL {$\alpha \in \lbrace0.1, 0.2, 0.3, 0.4\rbrace$}
            \STATE construct negative samples from $\mathcal{P}$ using Eq.~(\ref{equ:ng}) and insert them into $\mathcal{N}$.
            \STATE update training dataset: $\mathcal{D} = \mathcal{D} \cup \mathcal{N}$.
            \STATE update the parameters of $\mathcal{M}$ for $S$ epochs. In the second half of training process, compute error rate on $\mathcal{V}$ for each model, and insert this value at the end of $\mathcal{R}$.
            \FORALL {$I_{NS} \in \mathcal{N}$}
                \IF {$I_{NS}$ is misclassified}
                    \STATE remove corresponding pair from $\mathcal{P}$.
                \ENDIF
            \ENDFOR
            \STATE set $\mathcal{N} = \emptyset$.
            \STATE update current learning rate: $lr = lr \cdot 0.1$.
        \ENDFOR
        \STATE select $i$-th model which satisfies $\max \limits_{i} \lbrace r_i | r_i \in \mathcal{R}, r_i < \theta\rbrace$.
    \end{algorithmic}
    \caption{Enhanced training of CNN model with negative sample insertion}
    \label{alg:ng_insertion}
\end{algorithm}

\begin{figure}
\centering
\subfigure[$\mathcal{V}$]{
\begin{minipage}[b]{0.46\linewidth}
\includegraphics[width=1\linewidth]{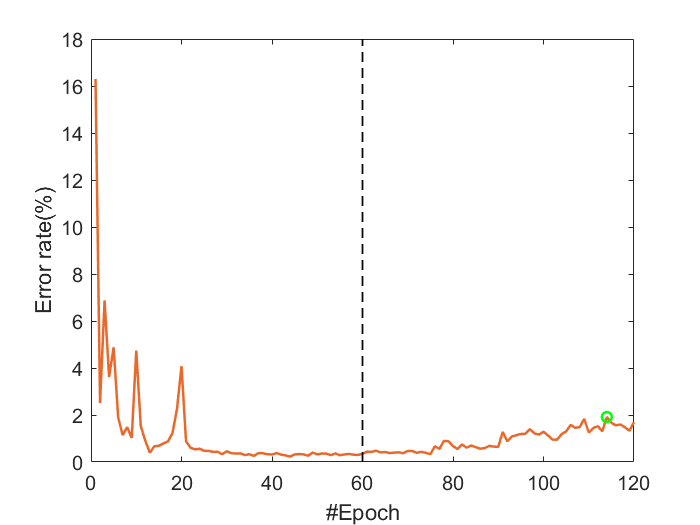}
\end{minipage}
}
\subfigure[Mb]{
\begin{minipage}[b]{0.46\linewidth}
\includegraphics[width=1\linewidth]{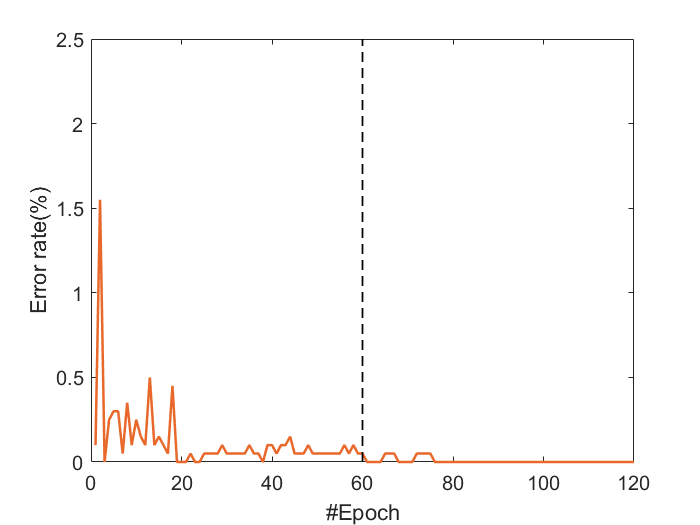}
\end{minipage}
}
\subfigure[Ma]{
\begin{minipage}[b]{0.46\linewidth}
\includegraphics[width=1\linewidth]{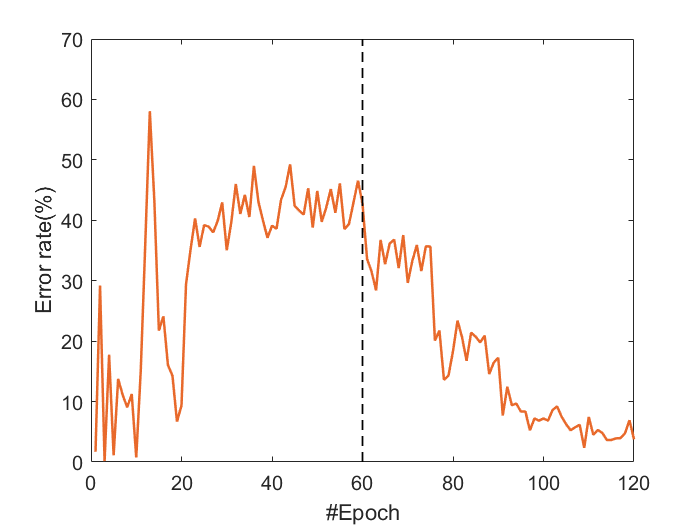}
\end{minipage}
}
\subfigure[Mc]{
\begin{minipage}[b]{0.46\linewidth}
\includegraphics[width=1\linewidth]{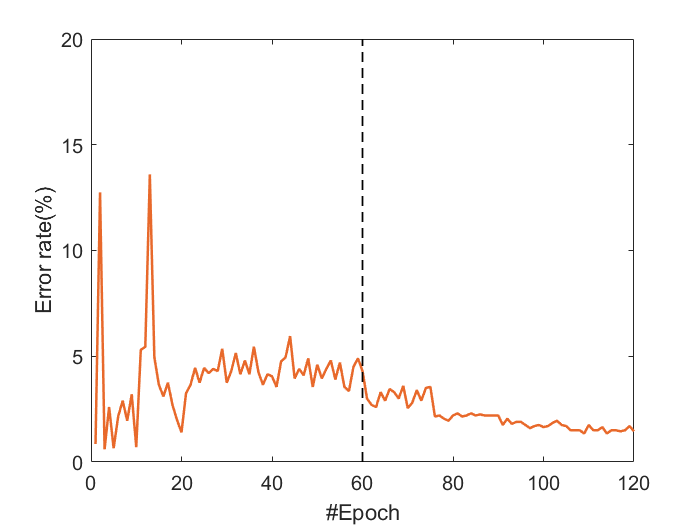}
\end{minipage}
}
\caption{Learning curves of a complete training of DecNet (with TanH in the first layer). The network is trained on Mb~\cite{zhang_colorful_2016}, and tested on Ma~\cite{larsson_learning_2016} and Mc~\cite{iizuka_let_2016}. The error rates (in $\%$) on CIs produced by these three methods are shown in (b), (c), and (d), respectively. The error rate on $\mathcal{V}$ is shown in (a). Black dotted line separates two training stages, where the first part uses the original training dataset for 60 epochs and the second part is the enhanced training with negative sample insertion (15 epochs for each insertion and in total 60 epochs). The green circle in (a) stands for the final selected model.
}
\label{fig:curve}
\end{figure}

The complete training process of proposed method includes two stages: (1) using the original training dataset to train the deep model from scratch until convergence; (2) iteratively adding new negative samples into the original training dataset and continuing to train the model as summarized in Algorithm~\ref{alg:ng_insertion}. Fig.~\ref{fig:curve} shows the learning curves of a complete training process. In the first stage, the error rates on $\mathcal{V}$ and CIs produced by Mb obviously decline in the first 20 epochs and the network reaches the stability after about 50 epochs, as shown in Fig.~\ref{fig:curve}(a) and (b). With the negative sample insertion, the error rate on $\mathcal{V}$ slightly increases, which can be found from the second part of Fig.~\ref{fig:curve}(a). However, the generalization performance of network has a significant improvement on CIs produced by Ma [Fig.~\ref{fig:curve}(c)] and a small improvement on Mc [Fig.~\ref{fig:curve}(d)]. More numerical and visual results (including t-SNE visualization after enhanced training) are given in Section~\ref{sec:experimental_results}.

\section{Experimental Results}
\label{sec:experimental_results}

\subsection{Implementation Details}
Our networks are implemented with PyTorch 0.3.1~\cite{pytroch}. The GPU version is GeForce\textsuperscript{\textregistered}~GTX 1080Ti of NVIDIA\textsuperscript{\textregistered}~corporation. All images in our experiments are resized to $256\times256$ using bicubic interpolation, and for each image, we rescale its pixel values to $[-1,1]$. Stochastic gradient descent (SGD) with a minibatch of 20 is used to train CNN models. Each minibatch contains 10 natural images and 10 colorized images. We randomly shuffle the order of training dataset after each epoch. For SGD optimizer, the momentum is 0.9 and the weight decay is 1e-4. The base learning rate is initialized to 1e-4. In our work, a complete training process includes two stages, respectively without and with negative sample insertion. In the first stage, we divide the learning rate by 10 every 20 epochs, and the training procedure stops after 60 epochs. In the second stage, the learning rate is continued to be divided by 10 every 15 epochs (it is enough to guarantee the convergence after new negative sample insertion), and the training procedure stops after 60 epochs, \textit{i.e.}, 4 iterations of negative sample insertion. For BN, we keep a running estimate of computed mean and variance in the training stage, and this running mean and variance is used for normalization in the testing stage~\cite{BN2015}. 

Following~\cite{guo_fake_2018}, we also employ the \emph{half total error rate} (HTER) to evaluate the performance of the proposed method. The HTER is defined as the average of misclassification rates (in $\%$) of NIs and CIs. In this work, all reported results of our method are the average of 7 runs.

\begin{table}
    \caption{The classification performance (HTER, in $\%$, lower is better) of different network architectures on three datasets constructed by Ma~\cite{larsson_learning_2016}, Mb~\cite{zhang_colorful_2016}, and Mc~\cite{iizuka_let_2016}, respectively. ``BaseNet+'' is an augmented version of ``BaseNet'' with more feature maps from the second to fourth layers (conv2-4). ``Arc.'' stands for ``Architecture''.}
    \label{table:activation}
    \setlength{\tabcolsep}{4.5pt} \centering
    \begin{tabular}{*{10}{c}}
        \toprule
        \multirow{2}{*}{Arc.} & \multicolumn{3}{c}{No activation} & \multicolumn{3}{c}{TanH} & \multicolumn{3}{c}{ReLU} \\
        \cmidrule(lr){2-4} \cmidrule(lr){5-7} \cmidrule(lr){8-10}
         & Ma & Mb & Mc & Ma & Mb & Mc & Ma & Mb & Mc \\
		\hline
        BaseNet & 0.66 & 0.32 & 0.87 & 0.56 & 0.19 & 0.72 & 0.63 & 0.26 & 0.77 \\
        \hline
        BaseNet+ & 0.69 & 0.33 & 0.87 & 0.58 & 0.24 & 0.69 & 0.69 & 0.27 & 0.78 \\
        \hline
        \textbf{DecNet} & 0.60 & 0.29 & 0.69 & 0.55 & 0.16 & 0.55 & 0.62 & 0.20 & 0.61 \\
        \bottomrule
    \end{tabular}
\end{table}

\begin{figure*}
\centering
\subfigure[Train on Ma, test on Mb and Mc]{
\begin{minipage}[b]{0.3\linewidth}
\includegraphics[width=1\linewidth]{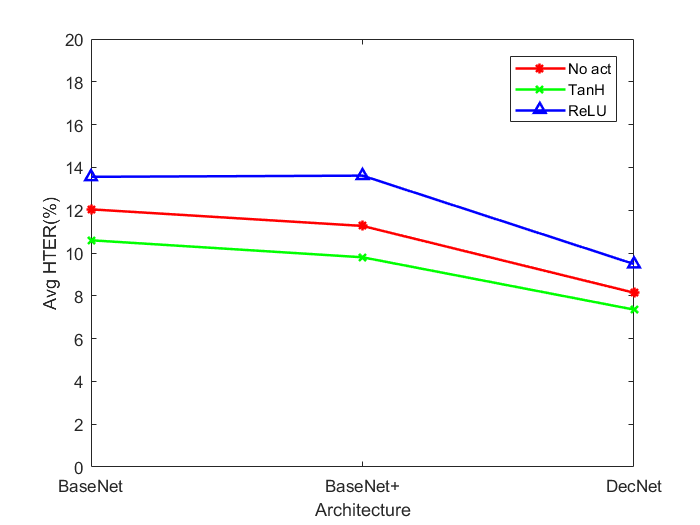}
\end{minipage}
}
\subfigure[Train on Mb, test on Ma and Mc]{
\begin{minipage}[b]{0.3\linewidth}
\includegraphics[width=1\linewidth]{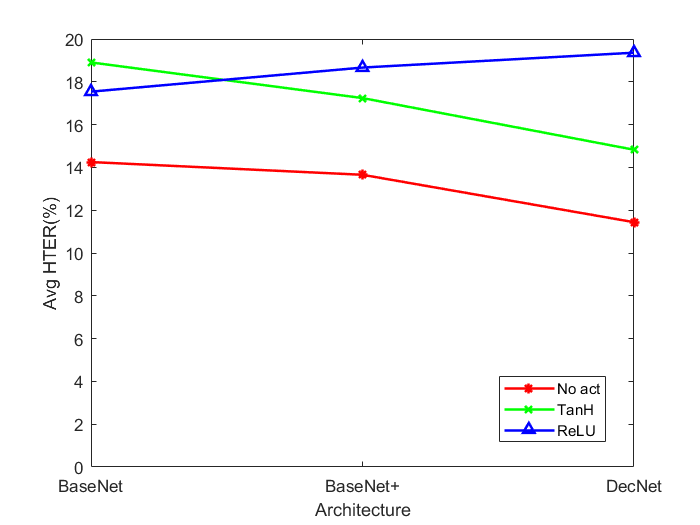}
\end{minipage}
}
\subfigure[Train on Mc, test on Ma and Mb]{
\begin{minipage}[b]{0.3\linewidth}
\includegraphics[width=1\linewidth]{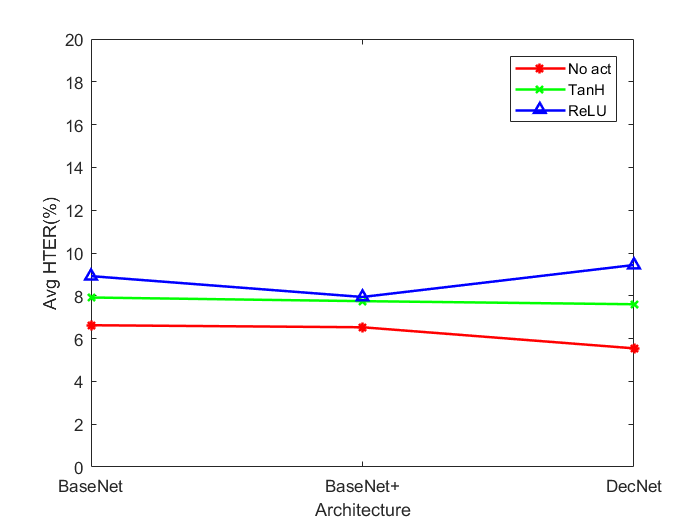}
\end{minipage}
}
\caption{The generalization performance (in HTER, lower is better) of different architectures on three different settings. From left to right, the CIs of training datasets are generated by Ma~\cite{larsson_learning_2016}, Mb~\cite{zhang_colorful_2016}, and Mc~\cite{iizuka_let_2016}, respectively. ``Avg HTER'' means average HTER of testing on datasets constructed by other two colorization methods. For example, (a) means training on dataset constructed by Ma, and testing on dataset constructed by Mb and Mc.} 
\label{fig:generalization_arc}
\end{figure*}

\subsection{Validation of Network Architecture Design}
\label{subsec:architecture}
Before evaluating the proposed method, we provide the details of datasets used in our experiments. Following~\cite{guo_fake_2018}, three state-of-the-art colorization algorithms, Ma~\cite{larsson_learning_2016}, Mb~\cite{zhang_colorful_2016}, and Mc~\cite{iizuka_let_2016} are adopted for producing CIs. NIs come from ImageNet dataset~\cite{ImageNet2009}. We use 10,000 natural images from ImageNet validation dataset to construct training dataset and validation dataset, and the ratio is 4:1. The exact indexes of these images are reported in~\cite{larsson_learning_2016}, and they are used for parameter selection and validation in~\cite{guo_fake_2018}. Then, we remove the 899 grayscale images and 1 CMKY (cyan, magenta, yellow, and black) image from the remaining 40,000 images of ImageNet validation dataset (the total number of images in this dataset is 50,000), and obtain 39,100 natural images to construct testing dataset. Note that, the magnitude of testing dataset is far larger than the setting reported in~\cite{guo_fake_2018}. We employ the three colorization methods mentioned above to produce the corresponding colorized images.

We first validate our network architecture design in terms of the classification performance and generalization capability of networks. As illustrated in Fig.~\ref{fig:network}, we first design a base network (BaseNet), which already has a good performance, and then improve our design by inserting a new branch into BaseNet, to obtain the final network (DecNet) that achieves further performance improvement. In order to verify that the performance improvement is not due to the increase of model parameters, we increase the number of feature maps from the second to fourth layers (conv2-4) of BaseNet (before: 64, 128, 256; after: 96, 192, 384) to obtain the augmented single-branch model BaseNet+. In total, the number of parameters of BaseNet, BaseNet+, and DecNet are 6.88M, 7.65M and 7.52M, respectively. Table~\ref{table:activation} reports the classification performance (in HTER, a measure of misclassification rate, so lower is better) of three network architectures on three different datasets (all with training and testing on the same colorization method), and each network considers three activation choices. We can find that all the three networks have very low misclassification rates, and that DecNet outperforms BaseNet and BaseNet+ for all the nine settings. In addition, all the three networks with TanH in the first layer have the best classification performance, and those without activation have the highest HTER. Two possible reasons are: (1) The non-linearity of TanH and ReLU help increase the approximation/learning capability of networks; (2) Different from ReLU, TanH keeps the sign of features which may provide useful information for classification of NIs and CIs.

\begin{table*}
    \caption{Performance of negative sample insertion of different network architectures on different datasets. Each network considers three activation types in the first layer. ``DecNet'' stands for network trained on original training dataset, and ``DecNet-i'' stands for network after enhanced training of the previously trained model (DecNet), with negative sample insertion. Starting from the second column, each consecutive two columns form a group. The former is the classification error rate (HTER, tested on the same colorization method as given in the second row), and the latter is the average generalization performance (Avg HTER, in italics, tested on the other two colorization methods). } 
    \label{table:ng}
    \setlength{\tabcolsep}{4.5pt} \centering
    \begin{tabular}{c|cc|cc|cc|cc|cc|cc|cc|cc|cc}
        \toprule
        \multirow{3}{*}{Arc.} & \multicolumn{6}{c|}{No activation} & \multicolumn{6}{c|}{TanH} & \multicolumn{6}{c}{ReLU} \\
        \cmidrule(lr){2-7} \cmidrule(lr){8-13} \cmidrule(lr){14-19}
         & \multicolumn{2}{|c|}{Ma} & \multicolumn{2}{c|}{Mb} & \multicolumn{2}{c|}{Mc} &
          \multicolumn{2}{c|}{Ma} & \multicolumn{2}{c|}{Mb} & \multicolumn{2}{c|}{Mc} & \multicolumn{2}{c|}{Ma} & \multicolumn{2}{c|}{Mb} & \multicolumn{2}{c}{Mc} \\
		\hline
        DecNet & 0.60 & \textit{8.15} & 0.29 & \textit{11.44} & 0.69 & \textit{5.55} & 0.55 & \textit{7.36} & 0.16 & \textit{14.83} & 0.55 & \textit{7.61} & 0.62 & \textit{9.49} & 0.20 & \textit{19.36} & 0.61 & \textit{9.44}\\
        \hline
        DecNet-i & 1.11 & \textit{5.24} & 0.96 & \textit{2.51} & 1.10 & \textit{2.29} & 1.03 & \textit{4.61} & 0.85 & \textit{3.01} & 0.98 & \textit{2.30} & 1.14 & \textit{6.23} & 0.92 & \textit{3.71} & 1.02 & \textit{3.47}\\
        \bottomrule
    \end{tabular}
\end{table*}

Fig.~\ref{fig:generalization_arc} shows the generalization performance of three network architectures when trained and tested on datasets constructed by different colorization algorithms. For each of the three test settings shown in Fig.~\ref{fig:generalization_arc}, when looking at all the nine combinations between three network architectures and three activation choices, it is always DecNet, combined with a certain activation choice, that has the lowest HTER, and recall that in the meanwhile DecNet has always the best classification performance, regardless of the activation type, when tested on the same colorization method (as shown in Table~\ref{table:activation}). If we check in Fig.~\ref{fig:generalization_arc} the performance separately for each activation choice, in general, DecNet has the best generalization performance except for two cases (out of nine) when combined with ReLU [(b) and (c)]; however ReLU is apparently not a suitable activation function in terms of generalization performance as shown in the figure. A possible explanation is that ReLU sets the output of some neurons of the first layer to be zero and thus destroys to some extent the extracted useful information. TanH and ``no activation'' have better generalization performance (with the latter slightly outperforming the former), implying that preservation of extracted information at first layer is helpful for generalizing better.

To summarize, our network DecNet can stably increase the classification accuracy regardless of activation in the first layer and improve the generalization performance in most cases (especially for TanH and ``no activation''). This improvement is attributed to the new branch, which can enhance the discrimination and variety of learned features. DecNet achieves a better trade-off than BaseNet+; the latter weakly improves the generalization performance but slightly decreases the classification accuracy. This implies that the performance improvement of CNN model is more dependent on suitable network architecture design, instead of simply increasing the number of feature maps. Concerning the activation type, TanH is considered a good choice, achieving a very satisfying compromise between classification accuracy and generalization. In contrast, although the classification performance of networks with ReLU in the first layer is good enough (though slightly worse than networks with TanH in the first layer, see Table~\ref{table:activation}), the generalization performance is the worst among three activation choices. This is probably due to the fact that ReLU destroys part of the initial information directly extracted from the input image, reflecting the importance of preserving richness of extracted features of the first layer. Networks without activation in the first layer have the lowest classification accuracy as shown in Table~\ref{table:activation}, but they can well preserve the extracted features at first layer, which then contributes to the good generalization performance (see Fig.~\ref{fig:generalization_arc}). Although the generalization capability of our network DecNet with different activation choices have small difference, this can be stably improved by our negative sample insertion method and the detailed results are given in the next subsection.

\subsection{Effect of Negative Sample Insertion}

In this paper, we propose negative sample insertion to further improve the generalization performance of our network. As described in Section~\ref{subsec:neg}, this enhanced training uses natural validation dataset $\mathcal{V}$ to select the final model, and we randomly select 20,000 NIs from ImageNet test dataset~\cite{ImageNet2009} to construct $\mathcal{V}$. Table~\ref{table:ng} reports the performance of our network before (the row of ``DecNet'') and after (the row of ``DecNet-i'') negative sample insertion\footnote{For the sake of clarity, generalization performance is presented in \textit{italics} in Table~\ref{table:ng}, and this is the same for subsequent tables.}. From Table~\ref{table:ng}, we can see that the effect of negative sample insertion, \textit{i.e.}, improving the generalization of network, is consistently stable for different activation choices. The negative sample insertion leads to slight decrease of the classification accuracy, however, the generalization performance of network usually has apparent improvement. For example, the initial generalization error of DecNet with ReLU trained on Mb is $19.36\%$, and then reduces to $3.71\%$ after enhanced training using negative samples, with a slight increase of classification error from $0.20\%$ to $0.92\%$. When the initial generalization error is relatively small, like $5.55\%$ of DecNet without any activation and trained on Mc, negative sample insertion still further decreases this value to $2.29\%$, while the classification error changes from $0.69\%$ to $1.10\%$. This is also consistent with previous analysis (Section~\ref{subsec:neg}) that there is a trade-off between the classification and the generalization performance, and our negative sample insertion method can achieve a satisfying trade-off.

\begin{figure*}
\centering
\subfigure[]{
\includegraphics[width=0.35\linewidth]{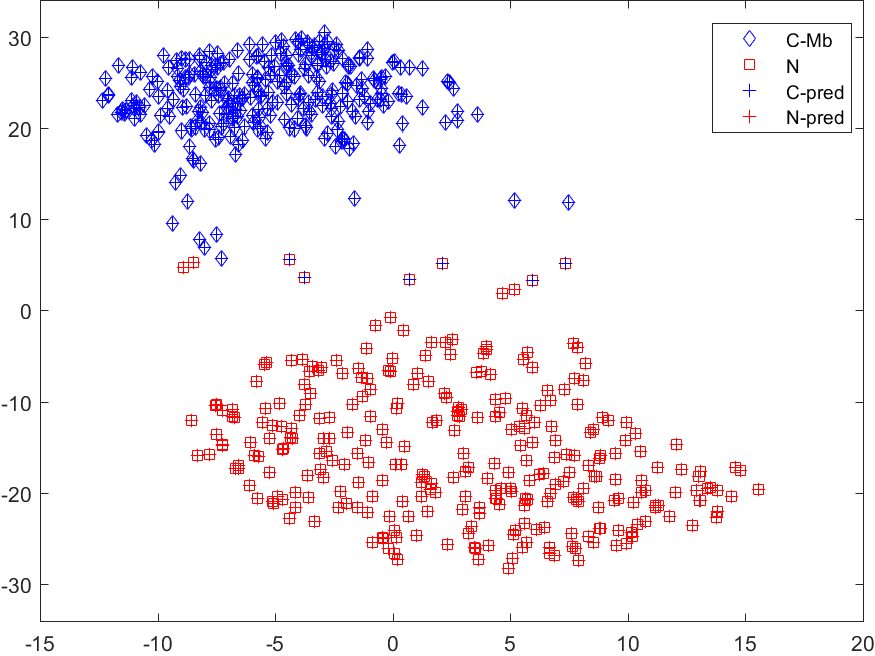}
}
\subfigure[]{
\includegraphics[width=0.35\linewidth]{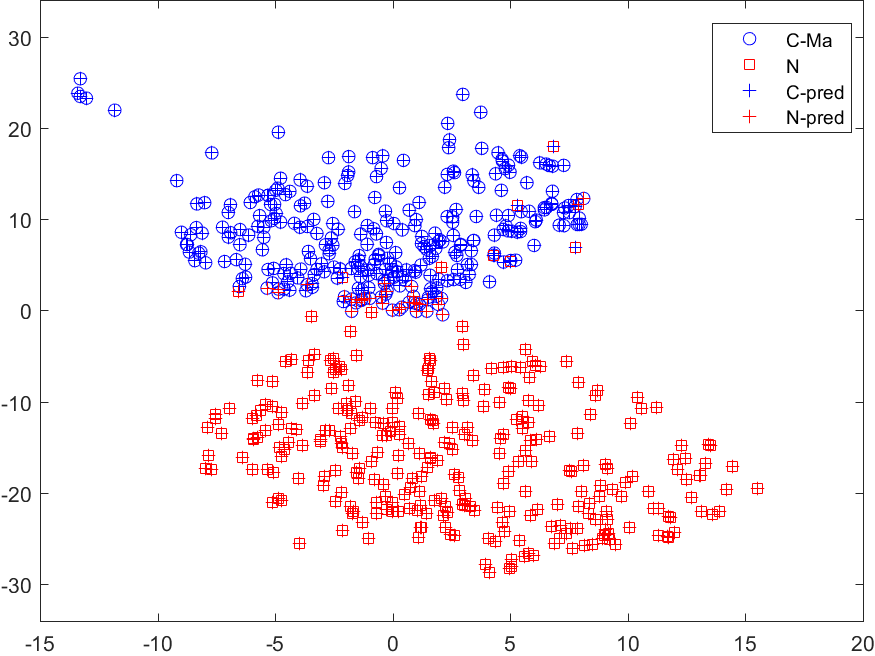}
}
\subfigure[]{
\includegraphics[width=0.35\linewidth]{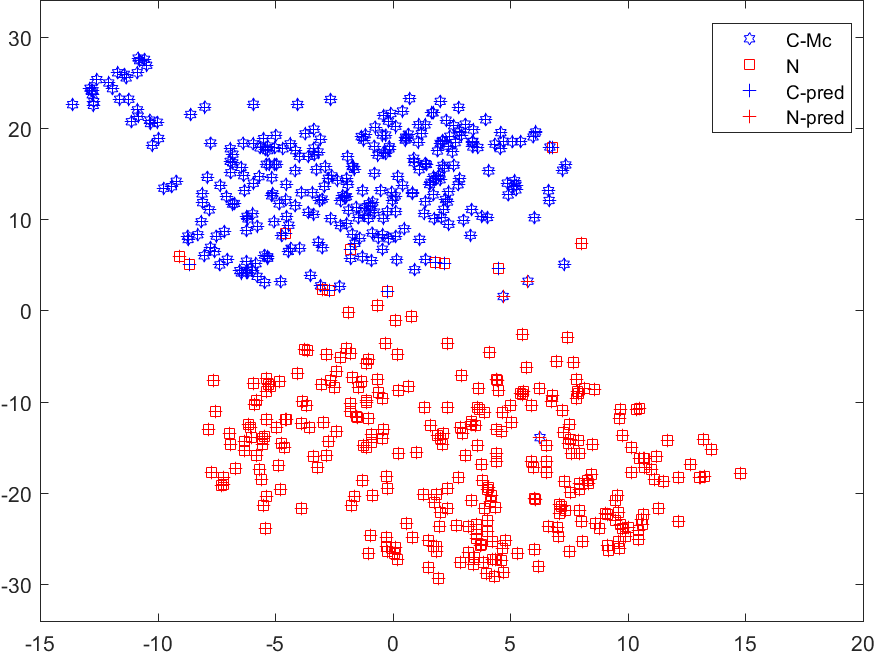}
}
\subfigure[]{
\includegraphics[width=0.35\linewidth]{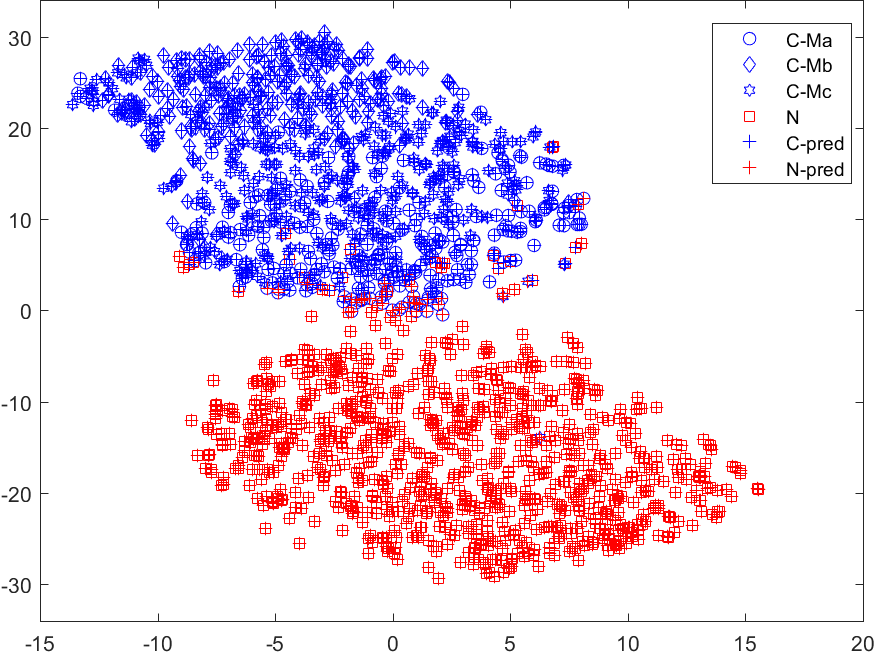}
}
\caption{The deep feature visualization of DecNet-i with t-SNE~\cite{TSNE2008}. The model is obtained through enhanced training of the previously trained model (used in Fig.~\ref{fig:boundary_o}) with negative sample insertion. The meaning of symbols is same as that of Fig.~\ref{fig:boundary_o}. It is worth noting that in t-SNE the transformation used for dimension reduction and the obtained visualization depend on the input data. Therefore, transformation and visualization in this figure are different from those of Fig.~\ref{fig:boundary_o}.}
\label{fig:boundary_ng}
\end{figure*}

In addition, we also visualize deep features of DecNet-i with TanH using t-SNE~\cite{TSNE2008}, and the results are shown in Fig.~\ref{fig:boundary_ng}. Here, deep features are the output of conv8 of DecNet-i, and its dimension is 512. The corresponding visualizations of the model before negative sample insertion are shown in Fig.~\ref{fig:boundary_o}. The testing data is also the same in Fig.~\ref{fig:boundary_ng} and Fig.~\ref{fig:boundary_o}. By comparing the border of correctly classified CIs, \textit{i.e.}, blue symbols with a blue + inside, in Fig.~\ref{fig:boundary_o}(d) and Fig.~\ref{fig:boundary_ng}(d), we can find that the latter has fewer misclassified CIs, and the classification boundary is pushed towards NIs. The CIs generated by ``unknown'' colorization algorithms, especially Ma~\cite{larsson_learning_2016}, are in consequence less misclassified, and this can be clearly observed by comparing Fig.~\ref{fig:boundary_o}(b) with Fig.~\ref{fig:boundary_ng}(b). This confirms that our negative sample insertion scheme can push the decision boundary towards NIs to some extent and then improve the generalization performance.

\subsection{Comparison with State-of-the-Art}
\label{subsec:state}
We experimentally compare the performance of our method (all networks with TanH in the first layer) with that of the state-of-the-art methods~\cite{guo_fake_2018}. We take the network with TanH as example for detailed comparison with \cite{guo_fake_2018}, for the sake of brevity. But our method with other activation types has in general consistently good performance; in particular the performance of the final network DecNet-i with different activations is very similar as shown in the last row of Table~\ref{table:ng}, and outperforms the methods in \cite{guo_fake_2018} as described below.

\begin{table}
    \caption{Comparison of the performance (HTER, in $\%$, lower is better) of our method with that of the state-of-the-art methods~\cite{guo_fake_2018} on ImageNet validation dataset~\cite{ImageNet2009}. ``FCID-HIST'' and ``FCID-FE'' are proposed in~\cite{guo_fake_2018}. Note that, the results of~\cite{guo_fake_2018} are obtained by testing on 2,000 images, and those of our method are from testing on 78,200 images. The generalization performance results are in italics.}
    \label{table:comp1}
    \setlength{\tabcolsep}{1.5pt} \centering
    \begin{tabular}{*{10}{c}}
        \toprule
        \multirow{2}{*}{Method} & \multicolumn{3}{c}{Ma} & \multicolumn{3}{c}{Mb} & \multicolumn{3}{c}{Mc} \\
        \cmidrule(lr){2-4} \cmidrule(lr){5-7} \cmidrule(lr){8-10}
         & Ma & Mb & Mc & Ma & Mb & Mc & Ma & Mb & Mc \\
        \hline
        FCID-HIST~\cite{guo_fake_2018} & 22.50 & \textit{28.00} & \textit{33.95} & \textit{26.95} & 24.45 & \textit{41.85} & \textit{38.15} & \textit{43.55} & 22.35 \\
        \hline
        FCID-FE~\cite{guo_fake_2018} & 22.30 & \textit{23.65} & \textit{31.70} & \textit{25.10} & 22.85 & \textit{34.25} & \textit{38.50} & \textit{36.15} & 17.30 \\
        \hline
				\hline
        BaseNet & 0.56 & \textit{10.57} & \textit{10.62} & \textit{31.65} & 0.19 & \textit{6.16} & \textit{13.93} & \textit{1.91} & 0.72 \\
        \hline
        DecNet & 0.55 & \textit{7.62} & \textit{7.09} & \textit{26.12} & 0.16 & \textit{3.53} & \textit{13.09} & \textit{2.12} & 0.55 \\
        \hline
        DecNet-i & 1.03 & \textit{5.09} & \textit{4.13} & \textit{4.41} & 0.85 & \textit{1.60} & \textit{2.83} & \textit{1.77} & 0.98 \\
        \bottomrule
    \end{tabular}
\end{table}

We first compare classification accuracy and generalization performance on ImageNet validation dataset~\cite{ImageNet2009}, and all testing results are shown in Table~\ref{table:comp1}. We can find that the classification accuracy (numbers not in italics in Table~\ref{table:comp1}) of our base architecture BaseNet is much improved compared with the two methods in \cite{guo_fake_2018} respectively denoted by FCID-HIST and FCID-FE, and the generalization performance (numbers in italics) is also better than that of two existing methods except for one case ($31.65\%$, \textit{i.e.}, training on Mb and testing on Ma). For DecNet (the second last row), the trend is almost same. There is a significant improvement of generalization performance through the negative sample insertion compared with \cite{guo_fake_2018}, and this can be observed by comparing the numbers in italics in same column of the rows of FCID-HIST, FCID-FE and DecNet-i of Table~\ref{table:comp1}. Furthermore, the results of~\cite{guo_fake_2018} are obtained by testing on 2,000 images (1,000 ImageNet images and corresponding 1,000 CIs) whose exact indexes remain unknown, and those of our method are testing on 78,200 images (39,100 ImageNet images and corresponding 39,100 CIs). In order to confirm the reliability and rationality of comparison, as an example, Table~\ref{table:multi} reports the statistical results of models trained on Ma (these models are same as those in the first group, \textit{i.e.}, from the second to fourth columns, in Table~\ref{table:comp1}). Practically, we run tests for 500 times and each time on 2,000 images (1,000 pairs of NIs and CIs) randomly selected from 78,200 images. We can see that the performance of our method is stably superior than that of \cite{guo_fake_2018} (comparing the three groups of results in Table~\ref{table:multi} and the first group of the rows of FCID-HIST and FCID-FE in Table~\ref{table:comp1}).

\begin{table}
    \caption{Multiple statistics of HTER of testing on 2,000 images (1,000 pairs of NIs and CIs) randomly selected from 78,200 images. The model is trained on Ma. We run 500 times and compute the maximum, mean, and minimum of HTER. The generalization performance results are in italics.}
    \label{table:multi}
    \setlength{\tabcolsep}{2.5pt} \centering
    \begin{tabular}{*{10}{c}}
        \toprule
        \multirow{2}{*}{Arc.} & \multicolumn{3}{c}{Maximum} & \multicolumn{3}{c}{Mean} & \multicolumn{3}{c}{Minimum}\\
        \cmidrule(lr){2-4} \cmidrule(lr){5-7} \cmidrule(lr){8-10}
         & Ma & Mb & Mc & Ma & Mb & Mc & Ma & Mb & Mc \\
        \hline
        BaseNet & 0.98 & \textit{12.19} & \textit{12.21} & 0.55 & \textit{10.56} & \textit{10.61} & 0.15 & \textit{9.02} & \textit{8.50} \\
        \hline
        DecNet & 1.08 & \textit{9.67} & \textit{8.16} & 0.55 & \textit{7.62} & \textit{7.08} & 0.14 & \textit{6.02} & \textit{5.57} \\
        \hline
        DecNet-i & 1.64 & \textit{6.81} & \textit{5.16} & 1.03 & \textit{5.06} & \textit{4.14} & 0.54 & \textit{3.75} & \textit{2.83} \\
        \bottomrule
    \end{tabular}
\end{table}

We then compare the performance of the cross-dataset test of our method and \cite{guo_fake_2018}, and the corresponding results are reported in Table~\ref{table:comp2}. The same as in~\cite{guo_fake_2018}, we also consider two cases: train on ImageNet validation dataset~\cite{ImageNet2009} and test on Oxford building dataset~\cite{Oxbuild2007} (called Oxbuild, which consists of 5,063 images); train on Oxbuild and test on ImageNet validation dataset. For each case, we consider three colorization methods: Ma~\cite{larsson_learning_2016}, Mb~\cite{zhang_colorful_2016}, and Mc~\cite{iizuka_let_2016}. The results of~\cite{guo_fake_2018} are obtained by testing on 2,000 images whose exact indexes remain unknown, and those of our method are from testing on 78,200 images for ImageNet validation dataset~\cite{ImageNet2009} and 10,126 images for Oxbuild~\cite{Oxbuild2007}. It can be observed from Table~\ref{table:comp2} that the classification performance of cross-dataset test of our method is much better than that of two methods proposed in \cite{guo_fake_2018}. In addition, the statistical results of multiple testings of our method on 2,000 images (like previous experiment of Table~\ref{table:multi}) have the consistently good appearance as well (for the sake of brevity we do not present these results here). A possible reason of good performance of our method on cross-dataset test is that the CNN model can to some extent find the essential difference between NIs and CIs from data and decrease the potential interference of image content.

\begin{table}
    \caption{Comparison of HTER (in $\%$, lower is better) of our method with that of the state-of-the-art methods~\cite{guo_fake_2018} (``FCID-HIST'' and ``FCID-FE'') on cross-dataset test. Note that, the results of~\cite{guo_fake_2018} are obtained by testing on 2,000 images, and those of our method are from testing on 78,200 images for ImageNet validation dataset~\cite{ImageNet2009} and 10,126 images for Oxbuild~\cite{Oxbuild2007}. ``ImageNet $\rightarrow$ Oxbuild'' means training on ImageNet validation dataset and testing on Oxbuild, and vise versa.}
    \label{table:comp2}
    \setlength{\tabcolsep}{4.5pt} \centering
    \begin{tabular}{*{10}{c}}
        \toprule
        \multirow{2}{*}{Dataset} & \multicolumn{3}{c}{ImageNet $\rightarrow$ Oxbuild} & \multicolumn{3}{c}{Oxbuild $\rightarrow$ ImageNet} \\
        \cmidrule(lr){2-4} \cmidrule(lr){5-7}
         & Ma & Mb & Mc & Ma & Mb & Mc \\
		\hline
        FCID-HIST~\cite{guo_fake_2018} & 22.85 & 21.50 & 30.95 & 43.45 & 30.75 & 36.60 \\
        \hline
        FCID-FE~\cite{guo_fake_2018} & 51.40 & 22.70 & 20.20 & 49.80 & 30.25 & 23.15  \\
        \hline
        DecNet & 1.15 & 0.11 & 1.73 & 2.04 & 1.88 & 2.85 \\
        \bottomrule
    \end{tabular}
\end{table}

\subsection{Qualitative Analysis and Misclassified Cases}
\begin{figure}
    \centering
    \includegraphics[width=0.8\linewidth]{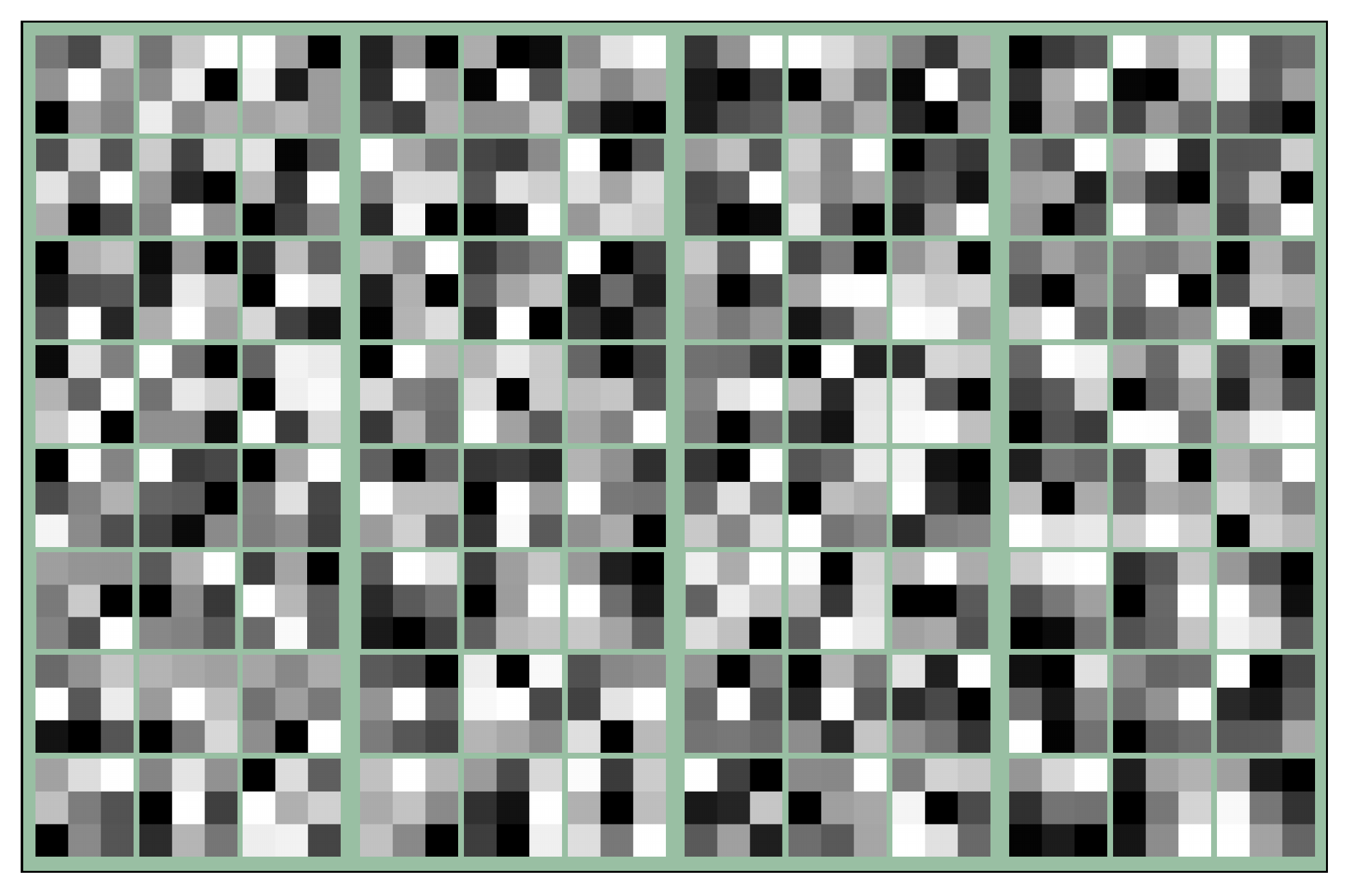}
    \caption{Visualization of the convolutional kernels of the first layer of our network. The filters are organized in groups of three (in columns) corresponding to the three color channels R, G and B. Brighter pixels stand for larger values. 
    }
    \label{fig:weight}
\end{figure}

In the following, we first conduct qualitative analysis through the visualization of the convolutional kernels and feature maps of well-trained models. Fig.~\ref{fig:weight} visualizes the convolutional kernels of the first layer of our network. For conventional computer vision classification tasks, the appearance of the first-layer filters of CNN models well trained on natural images is often common, in the sense that some of these filters are similar to Gabor filters and others resemble color blobs~\cite{AlexNet2012,zelier_visual_2014,yosinki_how_2014}. By contrast, the first-layer kernels of our method are almost all color sensitive and have no apparent orientation, which is reasonable for a network designed for detecting colorized images. In other words, the filters of the first layer of CNNs for computer vision tasks tend to take the image content as clues for image classification while our network rather considers the local color patterns as useful information for the classification of NIs and CIs.

\begin{figure*}
\centering
\includegraphics[width=0.9\linewidth]{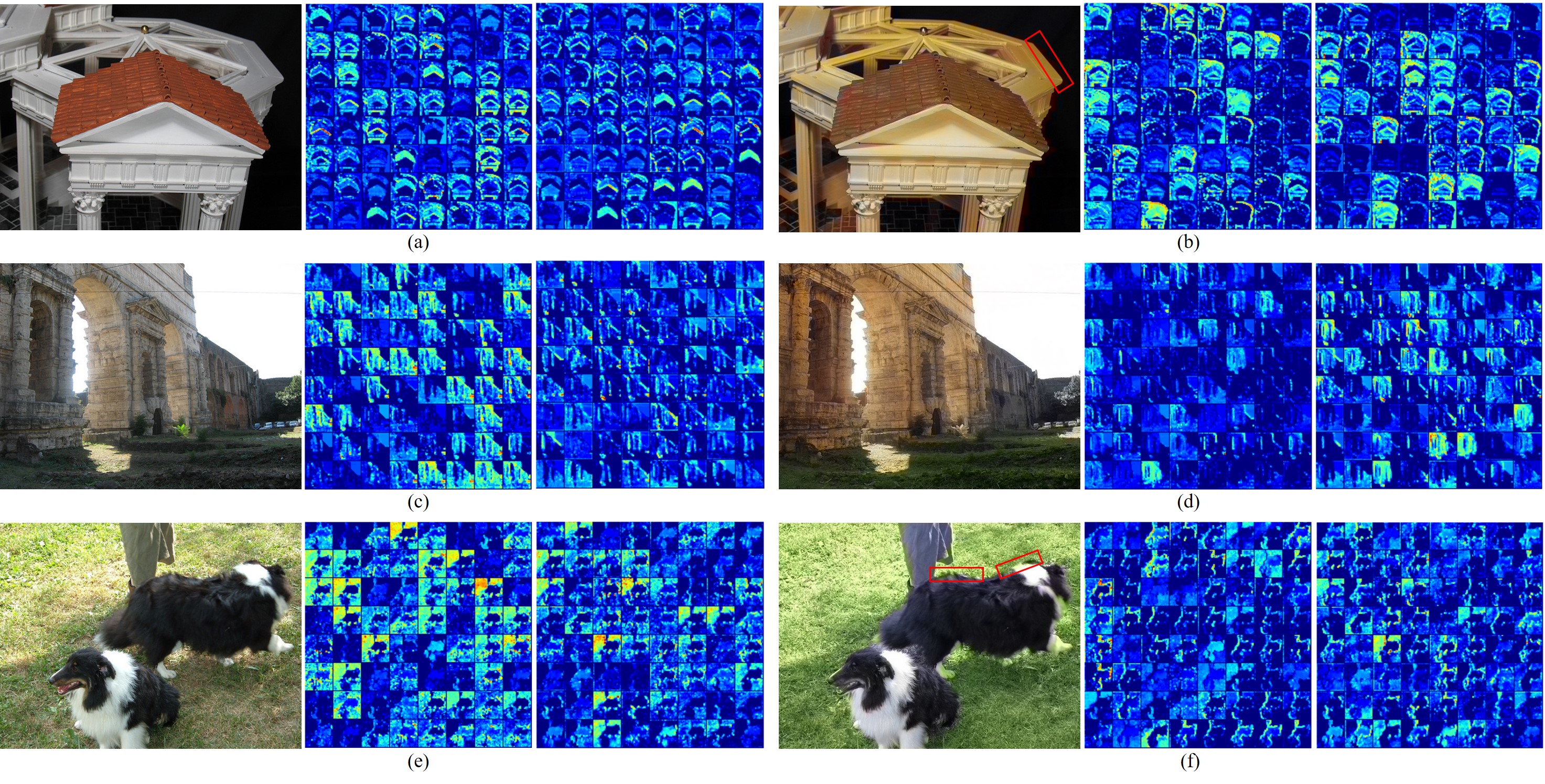}
\caption{Visualization of feature maps of conv4. We visualize the first 64 feature maps for each branch. The left groups of three columns correspond to NIs, and the right groups correspond to CIs. Each group consists of the RGB image (left), the feature map of the new branch (middle), and that of the base network (right). Hotter color stands for stronger activation. Red rectangle highlights the color bleeding region (best viewed with zooming on a big screen).}
\label{fig:feature}
\end{figure*}

The visualization of feature maps of CNN is often used to gain intuition of CNN models~\cite{zelier_visual_2014,deep_visual_toolbox}. Fig.~\ref{fig:feature} visualizes the first 64 feature maps of two branches at conv4. The left groups of three columns [(a), (c), and (e)] correspond to NIs and the right groups [(b), (d), and (f)] correspond to CIs. We can see that there is obvious difference between the feature maps of NIs and CIs for each branch, for example, Fig.~\ref{fig:feature}(a) and (b). In addition, the difference also exists between two branches of each test image. These differences imply that our network learns discriminative and enriched features through two-branch architecture. We observe that some feature maps in the middle and right of Fig.~\ref{fig:feature}(b) have strong response corresponding to the right region of CI in Fig.~\ref{fig:feature}(b) (\textit{i.e.}, the border of the building, and highlighted by red rectangle), where the CI has slight color bleeding. The similar phenomenon can also be found in the boundary region between grass and dog in Fig.~\ref{fig:feature}(f) (highlighted by two red rectangles), although here the color bleeding is to some extent masked by textures when compared with (b). For the middle and right of (e), \textit{i.e.}, the feature maps of two branches, high response in the grass is observed due to richness and naturalness of color, but low response in the counterparts of (f) is observed because the color is relatively monotonous in CI. Similarly, some feature maps of the new branch [the middle of Fig.~\ref{fig:feature}(c)] have high response for the gate and grass of NI in (c), whereas low response is found for (d). To summarize, these observations give the hint that our network can grab some useful clues, such as color bleeding and monotonous color, for classification of NIs and CIs.

\begin{figure}
    \centering
    \subfigure[]{
    \begin{minipage}[b]{0.95\linewidth}
    \centering
    \includegraphics[width=0.75\linewidth]{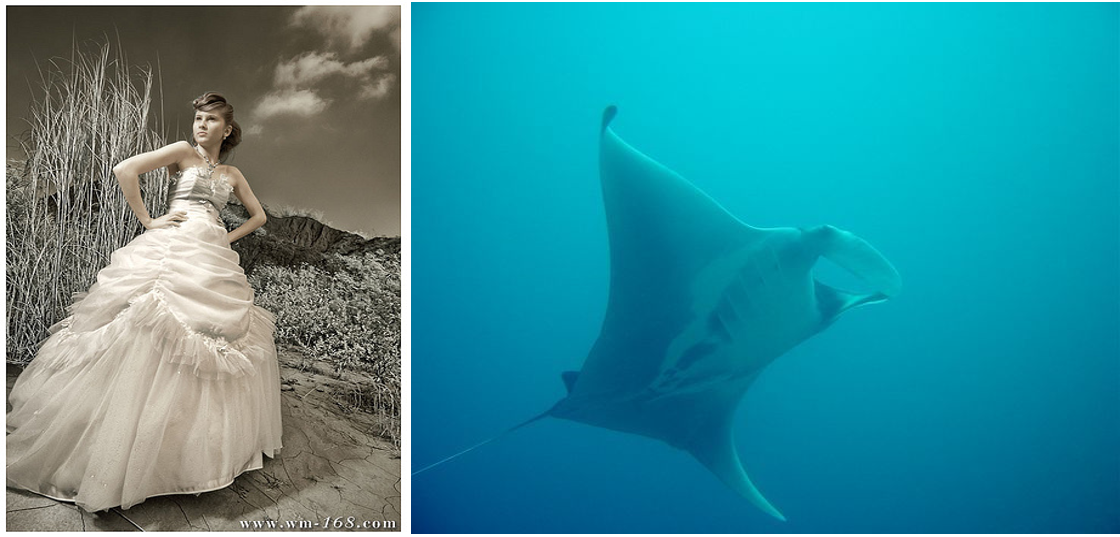}
    \end{minipage}
    }
    \subfigure[]{
    \begin{minipage}[b]{0.95\linewidth}
    \centering
    \includegraphics[width=0.75\linewidth]{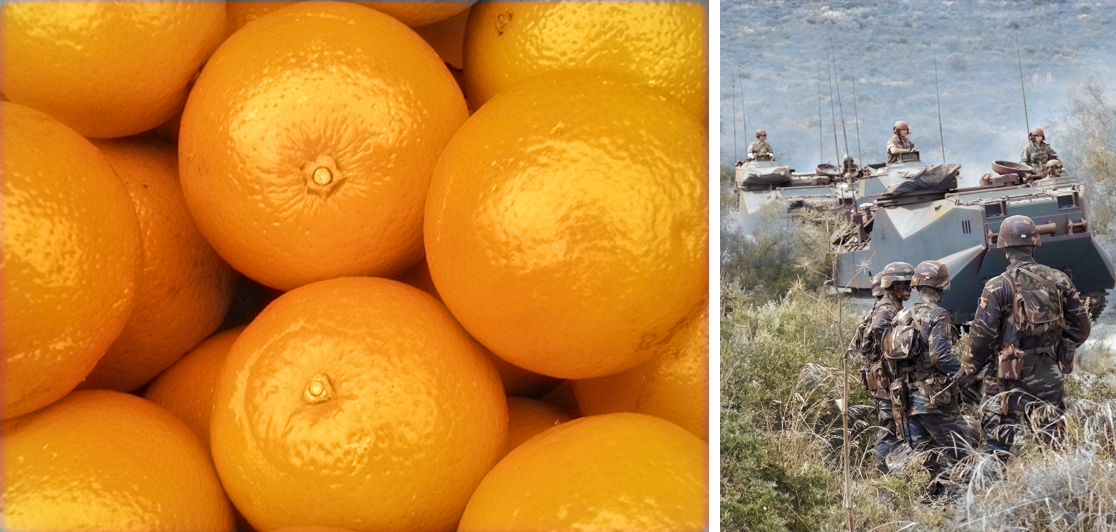}
    \end{minipage}
    }
    \caption{Misclassified cases. (a): NIs misclassified as CIs, and (b): CIs misclassified as NIs.
    }
    \label{fig:failure_examples}
\end{figure}

At last, several misclassified examples of our method are shown in Fig.~\ref{fig:failure_examples}. For (a), we can see that the color of first image is less saturated, and the color of the second image is sea blue for most of the area and somehow monotonous. Therefore, our network misclassifies these NIs as CIs. Conversely, the first image in (b) has relatively saturated color and clear boundary, and the second image has plausible and rich color, such as the grass and tank. These cues may have misled the classification decision of our network.

\section{Concluding Remarks}
\label{sec:conclusion}
In this paper, we proposed an ``end-to-end'' framework based on the convolutional neural network to distinguish between natural images and colorized images. We first designed a base CNN model, which outperformed state-of-the-art methods (in most cases) in terms of both classification and generalization performance. Afterwards, we designed and added a new branch to the base network, leading to a CNN with enhanced architecture and enriched features. This well-designed network not only improves the classification accuracy but also the generalization performance. Furthermore, we considered the challenging blind detection scenario and proposed an effective method based on negative sample insertion to further improve the generalization capability of our CNN model. Consequently, our network's generalization performance is obviously and stably improved while decreasing very slightly the classification accuracy. We plan to share the source code of our method to the research community.

In the future, we are willing to apply our two-branch network architecture to other image forensic problems, where enriched features may help improve the forensic performance, and also to optimize the architecture in a task adaptive manner. For example, we could optimize in a rigorous way some meta parameters including the number of feature maps (width) and layers (depth) of each stage of the CNN for each task. We also plan to employ the proposed negative-sample-based enhanced training to improve the generalization performance of other kinds of forensic methods whenever applicable. Concerning the classification of natural and colorized images or other related problems, it would be interesting to further improve the CNN architecture via modeling high-level semantic information or imitating the human perception process of the given task. Furthermore, an attractive research direction is to explore other approaches to understanding and enhancing the generalization performance of neural networks.

\section*{Acknowledgment}
We would like to thank Dr. G. Larsson, Dr. R. Zhang and Dr. S. Iizuka for kindly sharing the source code of their colorization algorithms respectively described in~\cite{larsson_learning_2016}, \cite{zhang_colorful_2016} and \cite{iizuka_let_2016}, Dr. L. van der Maaten for making the t-SNE tool~\cite{TSNE2008} publicly available, and Dr. Y. Guo for detailed and helpful discussions about their work~\cite{guo_fake_2018}.

%

\ifCLASSOPTIONcaptionsoff
  \newpage
\fi

\bibliographystyle{IEEEtran}
\bibliography{col}

\end{document}